%% file: main.tex
\documentclass{article} 
\usepackage{iclr2023_conference,times}

\PassOptionsToPackage{square,numbers}{natbib}

\usepackage[utf8]{inputenc} 
\usepackage[T1]{fontenc}    
\usepackage{hyperref}       
\usepackage{url}            
\usepackage{booktabs}       
\usepackage{amsfonts}       
\usepackage{nicefrac}       
\usepackage{microtype}      
\usepackage{xcolor}
\usepackage{booktabs}
\usepackage{graphics}
\usepackage{subcaption}
\usepackage{hyperref}
\usepackage{float}
\usepackage{multirow}
\floatstyle{plaintop}
\restylefloat{table}
\usepackage[pdftex]{graphicx}
\usepackage{pdflscape}
\usepackage{cancel}
\usepackage{amsmath}
\usepackage{dsfont}
\usepackage{placeins}
\usepackage{amsthm}
\hypersetup{
    colorlinks,
    linkcolor={red!50!black},
    citecolor={green!50!black},
    urlcolor={blue!80!black}
}
\usepackage{bm}
\hypersetup{pdftitle={Deep Variational Implicit Process}}

\newtheorem{definition}{Definition}

\title{Deep Variational Implicit Processes}

\author{
  Luis A. Ortega\textsuperscript{1},\quad Simón Rodríguez Santana\textsuperscript{2},\quad  Daniel Hernández-Lobato\textsuperscript{1} \\
  \textsuperscript{1}Universidad Autónoma de Madrid\quad \textsuperscript{2}ICMAT-CSIC\\
  {\small \texttt{\{luis.ortega,daniel.hernandez\}@uam.es}, \ \texttt{simon.rodriguez@icmat.es}}\\
}

\iclrfinalcopy
\begin{document}
\lineskip=0pt

\maketitle

\begin{abstract}
Implicit processes (IPs) are a generalization of Gaussian 
processes (GPs). IPs may lack a closed-form expression but 
are easy to sample from. Examples include, among others, 
Bayesian neural networks or neural samplers. IPs can be used 
as priors over functions, resulting in flexible models with 
well-calibrated prediction uncertainty estimates. Methods based 
on IPs usually carry out function-space approximate inference, 
which overcomes some of the difficulties of parameter-space 
approximate inference. Nevertheless, the approximations employed often 
limit the expressiveness of the final model, resulting, \emph{e.g.}, 
in a Gaussian predictive distribution, which can be restrictive. 
We propose here a multi-layer generalization of IPs called the 
Deep Variational Implicit process (DVIP). This generalization is 
similar to that of deep GPs over GPs, but it is more flexible due 
to the use of IPs as the prior distribution over the latent functions. 
We describe a scalable variational inference algorithm for 
training DVIP and show that it outperforms previous IP-based 
methods and also deep GPs. We support these claims via extensive 
regression and classification experiments. We also evaluate DVIP on 
large datasets with up to several million data instances to 
illustrate its good scalability and performance.
\end{abstract}

\section{Introduction}
\input{sections/introduction}

\section{Background} \label{sec:vip}
\input{sections/background}

\section{Deep variational implicit processes}
\input{sections/dvip}

\section{Related work}
\label{sec:related}
\input{sections/related}

\section{Experiments}
\input{sections/experiments}

\section{Discussion}
\input{sections/discussion}

\subsubsection*{Acknowledgments}
Authors gratefully acknowledge the use of the facilities of Centro de Computacion Cientifica (CCC) at Universidad Autónoma de Madrid. The authors also acknowledge financial support from Spanish Plan Nacional I+D+i, PID2019-106827GB-I00. Additional support was provided by the national project PID2021-124662OB-I00, funded by MCIN/ AEI /10.13039/501100011033/ and FEDER, "Una manera de hacer Europa", as well as project TED2021-131530B-I00, funded by MCIN/AEI /10.13039/501100011033 and by the European Union NextGenerationEU PRTR.

\bibliographystyle{abbrvnat}
\bibliography{refs}



\newpage
\appendix
\input{appendixes/appendix}

\end{document}

%% file: sections/introduction.tex
The Bayesian approach has become popular for capturing 
the uncertainty associated to the predictions made by models that otherwise provide 
point-wise estimates, such as neural networks (NNs) 
\citep{gelman2013bayesian, Gal2016Uncertainty, murphy2012machine}. 
However, when carrying out Bayesian inference, obtaining the posterior distribution 
in the space of parameters can become a limiting factor since it is often intractable.  
Symmetries and strong dependencies between parameters make the approximate inference problem much 
more complex. This is precisely the case in large deep NNs. Nevertheless, 
all these issues can be alleviated by carrying out approximate inference in 
the space of functions, which presents certain advantages due to 
the simplified problem. This makes the approximations obtained in this space 
more precise than those obtained in parameter-space, as shown in 
the literature \citep{ma2019variational, sun2019functional, santana2021sparse, ma2021functional}. 

\begin{figure}[t]
	\begin{center}
	\begin{tabular}{cc}
	\includegraphics[width=0.24\textwidth]{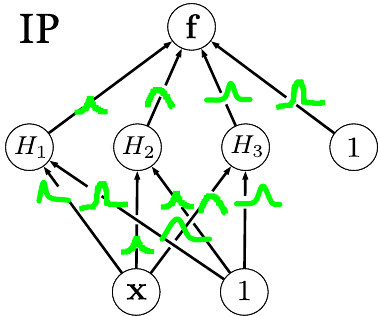} &
	\includegraphics[width=0.50\textwidth]{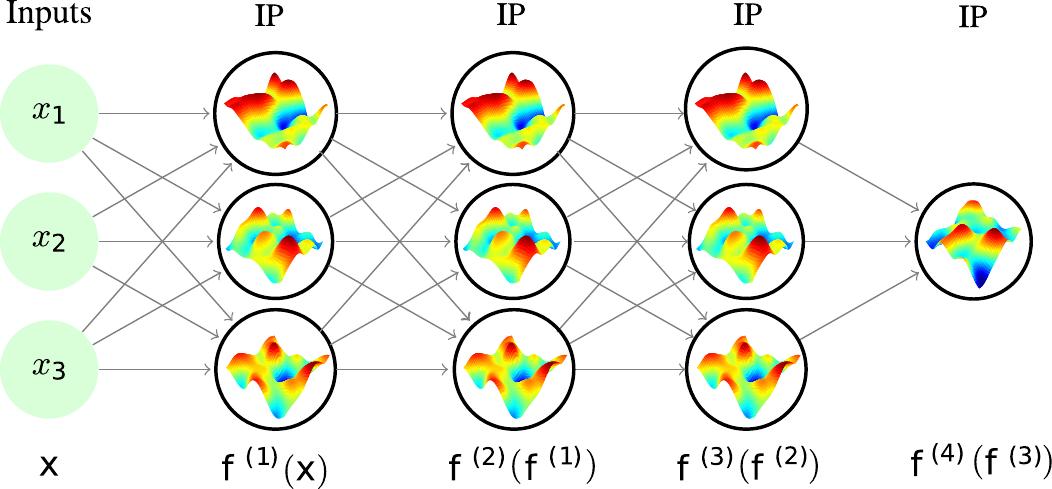} 
	\end{tabular}
	\end{center}
	\caption{(left) IP resulting from a BNN with random weights and biases 
	following a Gaussian distribution.
	A sample of the weights and biases generates a random function.
	(right) Deep VIP in which the input to an IP is the output of a previous IP. We consider 
	a fully connected architecture.}
	\label{fig:bnn_dvip}
\end{figure}

A recent method for 
function-space approximate inference is the \emph{Variational Implicit Process} (VIP) \citep{ma2019variational}. 
VIP considers an implicit process (IP) as the prior distribution over the target function.
IPs constitute a very flexible family of priors over functions that generalize 
Gaussian processes \citep{ma2019variational}. Specifically, IPs are processes that may lack 
a closed-form expression, but that are easy-to-sample-from. Examples include Bayesian neural 
networks (BNN), neural samplers and warped GPs, among others \citep{santana2021sparse}.
Figure \ref{fig:bnn_dvip} (left) shows a BNN, which is a particular case of an IP.
Nevertheless,  the posterior process of an IP is is intractable most of the times (except in the particular 
case of GPs). VIP addresses this issue by approximating the posterior using the 
posterior of a GP with the same mean and covariances as the prior IP. Thus, the approximation 
used in VIP results in a Gaussian predictive distribution, which may be too restrictive. 

Recently, the concatenation of random 
processes has been used to produce models of increased flexibility.
An example are \emph{deep GPs} (DGPs) 
in which a GP is used as the input of another GP, systematically \citep{damianou2013deep}. 
Based on the success of DGPs, it is natural 
to consider the concatenation of IPs to extend their capabilities 
in a similar fashion to DGPs. Therefore, we introduce in this paper deep 
VIPs (DVIPs), a multi-layer extension of VIP that provides
increased expressive power, enables more accurate predictions, gives better calibrated 
uncertainty estimates, and captures more complex patterns in the data.
Figure \ref{fig:bnn_dvip} (right) shows the architecture considered in 
DVIP. Each layer contains several IPs that are approximated using VIP.
Importantly, the flexibility of the IP-based prior formulation 
enables numerous models as the prior over functions, 
leveraging the benefits of, \emph{e.g.}, convolutional NNs, that 
increase the performance on image datasets. Critically, DVIP can adapt 
the prior IPs to the observed data, resulting in improved performance.
When GP priors are considered, DVIP is equivalent to a DGP. Thus, 
it can be seen as a generalization of DGPs. 

Approximate inference in DVIPs is done via variational inference (VI).
We achieve computational scalability in each unit using a 
linear approximation of the GP that approximates the 
prior IP, as in VIP \citep{ma2019variational}. The predictive distribution of
a VIP is Gaussian. However, since the inputs in the second and following layers
are random in DVIP, the final predictive distribution is non-Gaussian. 
This predictive distribution is intractable. Nevertheless, one can easily sample from 
it by propagating samples through the IP network shown in Figure \ref{fig:bnn_dvip} (right). 
This also enables a Monte Carlo approximation of the VI objective which 
can be optimized using stochastic techniques, as in 
DGPs \citep{salimbeni2017doubly}. Generating the required samples is 
straightforward given that the variational posterior depends only on the 
output of the the previous layers. This results in an iterative sampling procedure 
that can be conducted in an scalable manner. Importantly, the direct evaluation of 
covariances are not needed in DVIP, further reducing its cost compared to that of DGPs. 
The predictive distribution is a mixture of Gaussians (non-Gaussian), more flexible than that of VIP.

We evaluate DVIP in several experiments, both in 
regression and classification. They show that DVIP 
outperforms a single-layer VIP with a more complex IP 
prior. DVIP is also faster to train. We also show that DVIP gives results 
similar and often better than those of DGPs \citep{salimbeni2017doubly}, while 
having a lower cost and improved flexibility (due to the 
more general IP prior). Our experiments also show that 
adding more layers in DVIP does not over-fit and often improves results.

%% file: sections/background.tex
We introduce the needed background on IPs and the posterior approximation 
based on a linear model that will be used later on. First, consider the problem of inferring an unknown function 
\(f : \mathds{R}^M \to \mathds{R}\) given noisy observations \(\mathbf y = ( y_1, \dots,  y_N)^\text{T}\) at 
\(\mathbf X = (\mathbf x_1, \dots, \mathbf x_N)\). In the context 
of Bayesian inference, these observations are related to 
\(\mathbf f = (f(\mathbf {x}_1),\ldots,f(\mathbf{x}_N))^\text{T}\) via a likelihood, 
denoted as \(p( \mathbf y |\mathbf f)\). IPs
represent one of many ways to define a distribution over a function \citep{ma2019variational}.
\begin{definition}
An IP is a collection of random variables \(f (\cdot)\) such that any 
finite  collection
\(\mathbf f = \{ f(\mathbf x_1), f (\mathbf x_2), \dots , f(\mathbf x_N )\}\) is 
implicitly defined by the following generative process
\begin{align}
        \mathbf z & \sim p_{\mathbf z} (\mathbf z) \quad \mbox{and} 
\quad f (\mathbf x_n) = g_{\bm \theta} (\mathbf x_n, \mathbf z), \quad \forall n = 1, \dots, N.
\end{align} 
\end{definition}
An IP is denoted as \(f (\cdot) \sim \mathcal{IP} (g_{\bm \theta} (\cdot, \cdot), p_{\mathbf z} )\),
with $\bm \theta$ its parameters, $p_{\mathbf z}$ a source of noise, and $g_{\bm \theta} (\mathbf x_n, \mathbf z)$
a function that given $\mathbf{z}$ and $\mathbf{x}_n$ outputs $f(\mathbf x_n)$.
$g_{\bm \theta} (\mathbf x_n, \mathbf z)$ can be, \emph{e.g.}, a NN with weights specified 
by $\mathbf {z}$ and $\bm \theta$ using the reparametrization trick \citep{KingmaW13}. See Figure \ref{fig:bnn_dvip} (left).
Given \(\mathbf z \sim p_{\mathbf z}(\mathbf z)\) and 
\(\mathbf x\), it is straight-forward to generate a sample $f(\mathbf{x})$ using 
$g_{\bm \theta}$, \emph{i.e.}, \(f(\mathbf x) = g_{\bm \theta} (\mathbf x, \mathbf z)\). 

Consider an IP as the prior for an unknown function 
and a suitable likelihood \(p( \mathbf y | \mathbf f)\). In this 
context, both the prior \(p( \mathbf f | \mathbf X)\) and the 
posterior \(p(\mathbf f | \mathbf X, \mathbf y)\) are generally intractable, 
since the IP assumption does not allow for point-wise density estimation,
except in the case of a GP. To overcome this, 
in \cite{ma2019variational} the model's joint distribution,
\(p (\mathbf y, \mathbf f | \mathbf  X)\), is approximated as
\begin{equation}
    p (\mathbf y, \mathbf f | \mathbf X) \approx q(\mathbf y, \mathbf f | \mathbf X) = p(\mathbf  y | \mathbf f)q_{\mathcal{GP}}(\mathbf f | \mathbf X),
\end{equation}
where \(q_{\mathcal{GP}}\) is a Gaussian process with mean and covariance 
functions \(m(\cdot)\) and \(\mathcal{K}(\cdot, \cdot)\), respectively. These 
two functions are in turn defined by the mean and covariance functions of the prior IP,
\begin{align}
    m(\mathbf x) & = \mathds E [f(\mathbf x)], &
    \mathcal{K}(\mathbf x_1, \mathbf x_2) = & \mathds 
	E \left[ \left( f(\mathbf x_1) - m(\mathbf x_1)\right)\left( f(\mathbf x_2) - m(\mathbf x_2)\right) \right]\,,
\end{align}
which can be estimated empirically by sampling from \(\mbox{IP} (g_{\bm \theta} (\cdot, \cdot), p_{\mathbf z} )\) 
\citep{ma2019variational}.  Using $S$ Monte Carlo samples 
\(f_s(\cdot) \sim \mbox{IP} (g_{\bm \theta} (\cdot, \cdot), p_{\mathbf z} )\), \( s \in {1, \dots, S}\), 
the mean and covariance functions are
\begin{equation}\label{eq:prior_cov}
\begin{aligned}
    m^\star(\mathbf x)  &= \textstyle \frac 1 S  \sum_{s=1}^{S}  f_s(\mathbf x)\,, \quad \quad   \mathcal{K}^\star(\mathbf x_1, \mathbf x_2)
	 = \bm \phi(\mathbf x_1)^\text{T} \bm \phi(\mathbf x_2)\,, \\
    \bm \phi(\mathbf x_n) &= 1 / \sqrt{S}  \big(f_1(\mathbf x_n) - m^\star(\mathbf x_n), \ldots, f_S(\mathbf x_n) - m^\star(\mathbf x_n)\big)^\text{T}\,.
\end{aligned}
\end{equation}
Thus, the VIP's prior for $f$ is simply a GP approximating the prior IP, which can be, \emph{e.g.}, a BNN.
Critically, the samples $f_s(\mathbf x)$ keep the dependence w.r.t. the IP prior parameters $\bm \theta$, 
which enables prior adaptation to the observed data in VIP \citep{ma2019variational}.
Unfortunately, this formulation has the typical cost in $\mathcal{O}(N^3)$ of GPs \citep{rasmussen2005book}. 
To solve this and also allow for mini-batch training, the GP is approximated using a linear
model: $f(\mathbf{x}) = \bm \phi(\mathbf x)^\text{T} \mathbf{a} + m^\star(\mathbf x)$, 
where $\mathbf{a}\sim \mathcal{N}(\mathbf 0, \mathbf I)$.
Under this definition, the prior mean and covariances of $f(\mathbf{x})$ are given by (\ref{eq:prior_cov}).
The posterior of $\mathbf{a}$, $p(\mathbf{a}|\mathbf{y})$, is approximated using a Gaussian distribution, 
$q_{\bm{\omega}}(\mathbf a)=\mathcal{N}(\mathbf a|\mathbf m, \mathbf S)$, 
whose parameters $\bm \omega =\{\mathbf m,\mathbf S\}$ (and other model's parameters) are adjusted by maximizing 
the $\alpha$-energy
\begin{align}
	\mathcal{L}^\alpha(\bm \omega, \bm \theta, \sigma^2) & = \textstyle \sum_{n=1}^N 
	\frac{1}{\alpha}\log \mathds{E}_{q_\omega} \left[\mathcal{N}(y_n|f(\mathbf{x}_n), \sigma^2)^\alpha \right]
	- \text{KL}\big(q_\omega(\mathbf a) | p(\mathbf a)\big)\,,
	\label{eq:vip_objective}
\end{align}
where $\alpha = 0.5$, a value that provides good general results \citep{hernandez2016black, santana2022adversarial}.
A Gaussian likelihood is also assumed with variance \(\sigma^2\). 
Otherwise, $1$-dimensional quadrature methods are required to evaluate 
(\ref{eq:vip_objective}). Importantly, (\ref{eq:vip_objective}) allows for mini-batch training
to estimate $\bm \omega$, $\bm \theta$ and $\sigma^2$ from the data. Given, $q_{\bm{\omega}}(\mathbf a)$, it 
is straight-forward to make predictions. The predictive distribution for $y_n$ is, however,
limited to be Gaussian in regression.

%% file: sections/dvip.tex
Deep variational implicit processes (DVIPs) are models that consider
a deep implicit process as the prior for the latent function. A deep implicit process 
is a concatenation of multiple IPs, recursively defining an implicit prior over latent functions. 
Figure \ref{fig:bnn_dvip} (right) illustrates the architecture considered, which is fully connected.
The prior on the function at layer $l$ and unit $h$, \(f^l_h(\cdot)\), is an 
independent IP whose inputs are given by the outputs of the previous layer.
Let $H_l$ be the $l$-th layer dimensionality.
\begin{definition}
A deep implicit process is a collection of random variables 
$\{f^l_{h,n}: l=1,\ldots,L \land h=1\ldots,H_l \land n=1,\ldots,N\}$ 
such that each $f^l_{h,n}=f^l_h(\mathbf{f}^{l-1}_{\cdot,n})$, 
with $\mathbf{f}^{l-1}_{\cdot,n}$ the output of the previous layer in the network,
i.e., $\mathbf{f}^{l-1}_{\cdot,n} = (f^{l-1}_{1,n}, \ldots,f^{l-1}_{H_{l-1},n})^\text{T}$,
and each $f_h^l(\cdot)$ an independent IP:
$f^l_h (\cdot)  \sim \mathcal{IP}(g_{\bm \theta^l_h}(\cdot, \bm z), p_{\bm z})$,
where $\mathbf f^{0}_{\cdot,n} = \mathbf{x}_n$ symbolizes the initial input features to 
the network.
\end{definition}

As in VIP, we consider GP approximations for all the IPs in the deep IP prior 
defined above. These GPs are further approximated using a linear model, as in VIP.
This provides an expression for $f^{l}_{h,n}$ given the previous layer's output 
$\mathbf{f}^{l-1}_{\cdot,n}$ and $\mathbf{a}^l_h$, the coefficients of the 
linear model for the unit $h$ at layer $l$.
Namely, $f^l_{h,n} = \bm \phi_h^l(\mathbf{f}^{l-1}_{\cdot,n})^\text{T} \mathbf{a}^l_h +
m_{h,l}^\star(\mathbf{f}^{l-1}_{\cdot,n})$,
where $\bm \phi_h^l(\cdot)$ and $m_{h,l}^\star(\cdot)$ depend on the prior IP parameters $\bm{\theta}_h^l$.
To increase the flexibility of the model, we consider latent Gaussian 
noise around each $f^l_{h,n}$ with variance $\sigma^2_{l, h}$ (except for the last layer \(l = L\)). That is, $\bm \sigma^2_l = \{\sigma^2_{l, h}\}_{h=1}^{H_l}$ are the noise variances at layer $l$. 
Then, $p(f^l_{h,n}|\mathbf{f}^{l-1}_{\cdot,n},\mathbf{a}^l_h)$ is a Gaussian 
distribution with mean $\bm \phi_h^l(\mathbf{f}^{l-1}_{\cdot,n})^\text{T} \mathbf{a}^l_h +
m_{h,l}^\star(\mathbf{f}^{l-1}_{\cdot,n})$ and variance $\sigma^2_{l,h}$.
Let $\mathbf{A}^l = \{\mathbf{a}_1^l,\ldots,\mathbf{a}^l_{H_l}\}$
and $\mathbf{F}^l = \{\mathbf{f}^l_{\cdot,1},\ldots,\mathbf{f}_{\cdot,N}^l\}$. 
The joint distribution of all the variables (observed and latent) in DVIP is 
\begin{align}\label{eq:dvip_model}
    p\left(\mathbf y, \{\mathbf  F^l, \mathbf{A}^l\}_{l=1}^L \right) & =  \textstyle 
	\prod_{n=1}^N p(y_n |\mathbf f_{\cdot,n}^L) \textstyle 
	\prod_{l=1}^L \prod_{h=1}^{H_l} p(f^l_{h,n} |\mathbf{a}^l_h)p(\mathbf a^l_h)\,, 
\end{align}
where $p(\mathbf{a}^l_h) = \mathcal{N}(\mathbf{a}^l_h|\mathbf{0},\mathbf{I})$ and
we have omitted the dependence of $f^l_{h,n}$ on $\mathbf{f}^{l-1}_n$ to improve readability.
In (\ref{eq:dvip_model}), $\prod_{n=1}^N p(y_n | \mathbf{f}_{\cdot,n}^L)$ is the likelihood and
$\prod_{n=1}^N \prod_{l=1}^L \prod_{h=1}^{h_l} p(f^l_{h,n} |\mathbf{a}^l_h)p(\mathbf{a}^l_h)$ is the deep IP prior.
It may seem that the prior assumes independence across points. Dependencies are, however, obtained by 
marginalizing out each $\mathbf{a}_h^l$, which is tractable since the model is linear in $\mathbf{a}_h^l$.

We approximate the posterior $p\left(\{\mathbf  F^l, \mathbf{A}^l\}_{l=1}^L|\mathbf y  \right)$ using 
an approximation with a fixed and a tunable part, simplifying dependencies among layer units, but 
maintaining dependencies between layers:
\begin{align}
        q(\{\mathbf  F^l, \mathbf{A}^l\}_{l=1}^L) & = \textstyle 
	\prod_{n=1}^N \prod_{l=1}^L \prod_{h=1}^{H_l} p(f^l_{h,n} |\mathbf{a}^l_h)q(\mathbf a^l_h)\,, & q(\mathbf a^l_h) &= \mathcal{N}(\mathbf{a}^l_h|\mathbf{m}^l_h,\mathbf{S}_h^l)\,,
	\label{eq:approx_q_dvip}
\end{align}
where the factors $p(f^l_{h,n} |\mathbf{a}^l_h)$ are fixed to be the same factors as those
in (\ref{eq:dvip_model}), and the factors $q(\mathbf a^l_d)$ are the ones being specifically tuned. 
This approximation resembles that of \cite{salimbeni2017doubly} for DGPs, since the conditional distribution 
is fixed. However, we do not employ sparse GPs based on inducing points and use a linear model to 
approximate the posterior at each layer instead. Computing $q(\mathbf{f}^L_{\cdot,n})$ is intractable. However, 
one can easily sample from it, as described next.

Using (\ref{eq:approx_q_dvip}), we can derive a variational lower bound at whose maximum the
Kullback-Leibler (KL) divergence between 
$q(\{\mathbf  F^l, \mathbf{A}^l\}_{l=1}^L)$
and 
$p\left(\{\mathbf  F^l, \mathbf{A}^l\}_{l=1}^L|\mathbf y  \right)$
is minimized. Namely,
\begin{align}
\mathcal{L}\left(\Omega,\Theta, \{\bm \sigma_l^2\}_{l=1}^{L-1}\right) &
	= \textstyle \sum_{n=1}^N 
	\mathds{E}_q \big[ \log p\big(y_n|\mathbf{f}^L_{\cdot,n}\big)\big] -
	\sum_{l=1}^L \sum_{h=1}^{H_l} \text{KL}\big(q(\mathbf{a}_h^l) | p(\mathbf{a}_h^l)\big)\,,
	\label{eq:elbo_dvip_final}
\end{align}
where we have used the cancellation of factors,
and where $\Omega = \{\mathbf{m}_h^l,\mathbf{S}_h^l: l=1,\ldots,L \land h=1,\ldots,H_l\}$
are the parameters of $q$ and $\Theta = \{\bm \theta_h^l: l=1,\ldots,L \land h=1,\ldots,H_l\}$ 
are the DVIP prior parameters. Furthermore, $\text{KL}(\cdot|\cdot)$ denotes the KL-divergence
between distributions. $\text{KL}(q(\mathbf{a}_h^l)|p(\mathbf{a}_h^l))$ involves Gaussian
distributions and can be evaluated analytically. The expectations $\mathds{E}_q 
\left[ \log p(y_n|\mathbf{f}^L_{\cdot,n})\right]$ are intractable. 
However, they can be approximated via Monte Carlo, using the reparametrization trick 
\citep{KingmaW13}. Moreover,  $\sum_{n=1}^N \mathds{E}_q \left[ \log p(y_n|\mathbf{f}^L_{\cdot,n})\right]$ 
can be approximated using mini-batches.  Thus, (\ref{eq:elbo_dvip_final}) can be maximized 
w.r.t. $\Omega$, $\Theta$ and $\{\bm \sigma_l^2\}_{l=1}^{L-1}$, using stochastic optimization. 
Maximization w.r.t.  $\Theta$ allows for prior adaptation to the observed data, which is a 
key factor when considering IP priors. Appendix \ref{sec:app_elbo} has all the details about the 
derivation of (\ref{eq:elbo_dvip_final}). 

\paragraph{Sampling from the marginal posterior approximation.}
The evaluation of (\ref{eq:elbo_dvip_final})  requires samples from $q(\mathbf{f}^L_{\cdot,n})$
for all the instances in a mini-batch. This marginal only depends on the variables 
of the inner layers and units $f_{h,n}^l$ corresponding to the $n$-th instance.
See Appendix \ref{sec:app_marginals}.
Thus, we can sample from $q(\mathbf{f}^L_{\cdot,n})$ by recursively propagating samples 
from the first to the last layer, using $\mathbf{x}_n$ as the input. 
Specifically, $p(f^l_{h,n}|\mathbf{f}^{l-1}_{\cdot,n},\mathbf{a}^l_h)$ is Gaussian with a linear
mean in terms of $\mathbf{a}_h^l$, and $q(\mathbf{a}_h^l)$ is Gaussian. Thus,
$q(f^l_{h,n}|\mathbf{f}^{l-1}_{\cdot,n})=\int p(f^l_{h,n}|\mathbf{f}^{l-1}_{\cdot,n},
\mathbf{a}^l_h) q(\mathbf{a}_h^l) \, \text{d} \mathbf{a}_h^l$  is also Gaussian
with mean and variance:
\begin{align}\label{eq:pred_gauss_layer}
        \hat{m}^l_{h,n}(\mathbf{f}^{l-1}_{\cdot,n}) &= \bm{\phi}_h^l(\mathbf{f}^{l-1}_{\cdot,n})^\text{T} \mathbf{m}^l_h + m_{h,l}^\star(\mathbf{f}^{l-1}_{\cdot,n})\,, &
        \hat{v}^l_{h,n}(\mathbf{f}^{l-1}_{\cdot,n}) &=  \bm{\phi}_h^l(\mathbf{f}^{l-1}_{\cdot,n})^\text{T} \mathbf{S}^l_h \bm{\phi}_h^l(\mathbf{f}^{l-1}_{\cdot,n}) +
    	\sigma^2_{l,h}\,,
\end{align}
where $\sigma^2_{L,h} = 0$ and $\mathbf{m}^l_h $ and $\mathbf{S}^l_h$ are the parameters 
of $q(\mathbf{a}_h^l)$. Initially, let $l=1$. We set $\hat{\mathbf{f}}^0_{\cdot,n}=\mathbf{x}_n$ and generate a sample 
from $q(f^l_{h,n}|\hat{\mathbf{f}}^{0}_{\cdot,n})$ for $h=1,\ldots,H_l$.
Let $\hat{\mathbf{f}}^{l}_{\cdot,n}$ be that sample. Then,
we use $\hat{\mathbf{f}}^{l}_{\cdot,n}$ as the input for the next layer. This process
repeats for $l=2,\ldots,L$, until we obtain $\hat{\mathbf{f}}^{L}_{\cdot,n} \sim q(\mathbf{f}^L_{\cdot,n})$.

\paragraph{Making predictions for new instances.} Let $\mathbf{f}_{\cdot,\star}^L$ be the values
at the last layer for the new instance $\mathbf{x}_\star$. We approximate $q(\mathbf{f}_{\cdot,\star}^L)$
by propagating $R$ Monte Carlo samples through the network. Then,
\begin{align}\label{eq:pred}
    q(\mathbf{f}_{\cdot,\star}^L)  & \approx R^{-1}
	\textstyle \sum_{r=1}^R \prod_{h=1}^{H_L}
	\mathcal{N}(f_{h,\star}^L|\hat{m}_{h,\star}^L(\hat{\mathbf{f}}_{\cdot,\star}^{L-1,r}), 
	\hat{v}_{h,\star}^L(\hat{\mathbf{f}}_{\cdot,\star}^{L-1,r}))\,,
\end{align}
where $\hat{\mathbf{f}}_{\cdot,\star}^{L-1,r}$ is the $r$-th sample arriving at layer $L-1$
and $\hat{m}_{h,\star}^L(\cdot)$ and $\hat{v}_{h,\star}^L(\cdot)$ are given by (\ref{eq:pred_gauss_layer}).
We note that (\ref{eq:pred}) is a Gaussian mixture, which is expected to be more flexible than
the Gaussian predictive distribution of VIP. We set $R$ to $100$ for testing and
to $1$ for training, respectively. Computing, $p(y_\star) = \mathds{E}_q[p(y_\star|\mathbf{f}_{\cdot,\star}^L)]$ 
is tractable in regression, and can be approximated using $1$-dimensional quadrature in binary 
and multi-class classification, as in DGPs \citep{salimbeni2017doubly}.

\paragraph{Input propagation.} Inspired by the \emph{skip layer} approach of, 
\emph{e.g.} ResNet \citep{he2016identity}, and the addition 
of the original input to each layer in DGPs \citep{duvenaud2014avoiding, salimbeni2017doubly}, 
we implement the same approach here. For this, we add the previous input to the mean
in (\ref{eq:pred_gauss_layer}) if the input and output dimension of the layer is the same, 
except in the last layer, where the added mean is zero.  Namely, $\hat{m}^l_{h,n}(\mathbf{f}^{l-1}_{\cdot,n}) = 
\bm{\phi}_h^l(\mathbf{f}^{l-1}_{\cdot,n})^\text{T} \mathbf{m}^l_h + m_{h,l}^\star(\mathbf{f}^{l-1}_{\cdot,n}) + f_{h,n}^{l-1}$,
for $l=1,\ldots,L-1$.

\paragraph{Computational cost.} The cost at layer $l$ in DVIP is
in $\mathcal{O}(BS^2H_l)$, if $B > S$, with \(S\) the number of samples from the prior IPs, \(B\) 
the size of the mini-batch and \(H_l\) the output dimension of the layer. This cost is similar to 
that of a DGP, which has a squared cost in terms of $M$, the number of inducing points 
\citep{hensman2013gaussian,bui2016deep,salimbeni2017doubly}. In \cite{salimbeni2017doubly},
the cost at each layer is \(\mathcal{O}(BM^2H_l)\), if $B > M$. In our work, however, the number of 
prior IP samples $S$ is smaller than the typical number of inducing points in DGPs. 
In our experiments we use $S = 20$, as suggested for VIP \citep{ma2019variational}. 
Considering a DVIP with $L$ layers, the total cost is $\mathcal{O}(BS^2(H_1 + \dots + H_L))$. 
Our experiments show that, despite the generation of the prior IP samples, DVIP is faster than
DGP, and the gap becomes bigger as $L$ increases.

%% file: sections/related.tex
The relation between GPs and IPs has been previously studied. 
A 1-layer BNN with cosine activations and infinite width is equivalent to a GP 
with RBF kernel \citep{hensman2017variational}. A deep BNN is equivalent to a GP 
with a compositional kernel \citep{cho2009kernel}, as shown by \cite{lee2017deep}. 
These methods make possible to create expressive kernels for GPs. 
An inverse reasoning is used by \cite{flam2017mapping}, where 
GP prior properties are encoded into the prior weights of a BNN. 
\cite{fortuin2022priors} provides an extensive review about prior 
distributions in function-space defined by BNNs and stochastic processes. 
They give methods of learning priors for these models from data.

VIP \citep{ma2019variational} arises from the 
treatment of BNNs as instances of IPs. For this, an approximate GP is used 
to assist inference. Specifically, a prior GP is built with mean and covariance function
given by the prior IP, a BNN.  VIP can make use of the more flexible IP prior, whose
parameters can be inferred from the data, 
improving results over GPs \citep{ma2019variational}. However, 
VIP's predictive distribution is Gaussian.
DVIP overcomes this problem providing a non-Gaussian predictive distribution.
Thus, it is expected to lead to a more flexible model with better calibrated uncertainty estimates. 
DVIP differs from deep kernel learning (DKL) \citep{wilson2016deep, wilson2016stochastic},
where a GP is applied to a non-linear transformation of the inputs. Its predictive distribution 
is Gaussian, unlike that of DVIP and the non-linear transformation of DKL 
ignores epistemic uncertainty, unlike IPs.

There are other methods that have tried to make inference using IPs.
\cite{sun2019functional} propose the \emph{functional Bayesian neural networks} (fBNN), 
where a second IP is used to approximate the posterior of the first IP. This is a 
more flexible approximation than that of VIP. However, because both the prior and 
the posterior are implicit, the noisy gradient of the variational ELBO is intractable and 
has to be approximated. For this, a spectral gradient estimator is used \citep{shi2018spectral}. 
To ensure that the posterior IP resembles the prior IP in data-free regions,
fBNN relies on uniformly covering the input space. In high-dimensional spaces this can lead 
to poor results. Moreover, because of the spectral gradient estimator fBNN cannot tune the 
prior IP parameters to the data. In the particular case of a GP prior, fBNN simply maximizes 
the marginal likelihood of the GP w.r.t. the prior parameters. However, a GP 
prior implies a GP posterior. This questions using a second IP for posterior approximation. Recent works have employed a first order Taylor approximation to linearize the IP, approximating their implicit distribution by a GP \citep{rudner2021tractable, immer2021improving}. This leads to another GP approximation to an IP, different of VIP, where the computational bottleneck is located at computing the Jacobian of the linearized transformation instead of samples from the prior. More recent work in parameter-space considers using a repulsive term in BNN ensembles to guarantee the diversity among the members, avoiding their collapse in the parameter space \citep{d2021repulsive}. 

Sparse implicit processes (SIPs) use inducing points for approximate 
inference in the context of IPs \citep{santana2021sparse}. SIP does not have the 
limitations of neither VIP nor fBNN. It produces flexible predictive distributions 
(Gaussian mixtures) and it can adjust its prior parameters to the data. SIP, however, relies on 
a classifier to estimate the KL-term in the variational ELBO, which adds computational cost. 
SIP's improvements over VIP are orthogonal to those of DVIP over VIP and, in principle, SIP 
may also be used as the building blocks of DVIP, leading to even better results.

Functional variational inference (FVI) minimizes the KL-divergence between stochastic 
process for approximate inference \citep{ma2021functional}. Specifically, 
between the model's IP posterior and a second IP, as in fBNN. This is done efficiently by
approximating first the IP prior using a stochastic process generator (SPG). Then, 
a second SPG is used to efficiently approximate the posterior of the previous SPG.
Both SPGs share key features that make this task easy. However, FVI is also limited, as 
fBNN, since it cannot adjust the prior to the data. This questions its practical utility. 

As shown by \cite{santana2021sparse}, adjusting the prior IP to the observed data 
is key for accurate predictions. This discourages using fBNN and FVI 
as building blocks of a model using deep IP priors on the target function.
Moreover, these methods do not consider deep architectures such as the one 
in Figure \ref{fig:bnn_dvip} (right). Therefore, we focus on comparing with VIP, as DVIP generalizes VIP.

To our knowledge, the concatenation of IPs with the goal of describing priors over 
functions has not been studied previously. However, the concatenation of GPs 
resulting in deep GPs (DGPs), has received a lot of attention 
\citep{lawrence2007hierarchical,bui2016deep,cutajar2016random,salimbeni2017doubly,havasi2018,yu2019implicit}. 
In principle, DVIP is a generalization of DGPs in the same way as IPs generalize GPs. Namely, the
IP prior of each layer's unit can simply be a GP. Samples from such a GP prior can be efficiently 
obtained using, \emph{e.g.}, a 1-layer BNN with cosine activation functions that is wide enough 
\citep{rahimi2007,cutajar2016random}. 

The posterior approximation used in DVIP is similar to that used in the context
of DGPs by \cite{salimbeni2017doubly}. Moreover, each prior IP is approximated 
by a GP in DVIP. However, in spite of these similarities, there are important differences
between our work and that of \cite{salimbeni2017doubly}. Specifically, instead of relying 
on sparse GPs to approximate each GP within the GP network, DVIP uses a linear
GP approximation that needs no specification of inducing points. 
Furthermore, the covariance function used in DVIP is more flexible and specified by the
assumed IP prior. DGPs are, on the other hand, restricted to GP priors with specific covariance 
functions. DVIP has the extra flexibility of considering a wider range of IP priors that need not 
be GPs. Critically, in our experiments, DVIP significantly outperforms DGPs in image-related datasets, 
where using specific IP priors based, \emph{e.g.}, on convolutional neural networks, 
can give a significant advantage over standard DGPs. Our experiments also show that DVIP is faster 
to train than the DGP of \cite{salimbeni2017doubly}, and the difference becomes larger 
as the number of layers $L$ increases.

%% file: sections/experiments.tex
\defcitealias{salimbeni2017doubly}{DS-DGP}

We evaluate the proposed method, DVIP, on
several tasks. We 
use \(S = 20\) and a BNN as the IP prior for each unit. 
These BNNs have 2 layers of \(10\) units each 
with \emph{tanh} activations, as in \cite{ma2019variational}. 
We compare DVIP with VIP \citep{ma2019variational} and DGPs, closely following 
\cite{salimbeni2017doubly}. We do not compare results with fBNN nor FVI, described in Section 
\ref{sec:related}, because they cannot tune the prior IP parameters 
to the data nor they do consider deep architectures as the one in Figure \ref{fig:bnn_dvip} (right). 
An efficient PyTorch implementation of DVIP is found in the supplementary material. 
Appendix \ref{sec:app_exp_settings} has all the details about the experimental 
settings considered for each method.

\begin{figure}[htb!]
    \centering
    \makebox[\textwidth]{\includegraphics[width=0.9\textwidth]{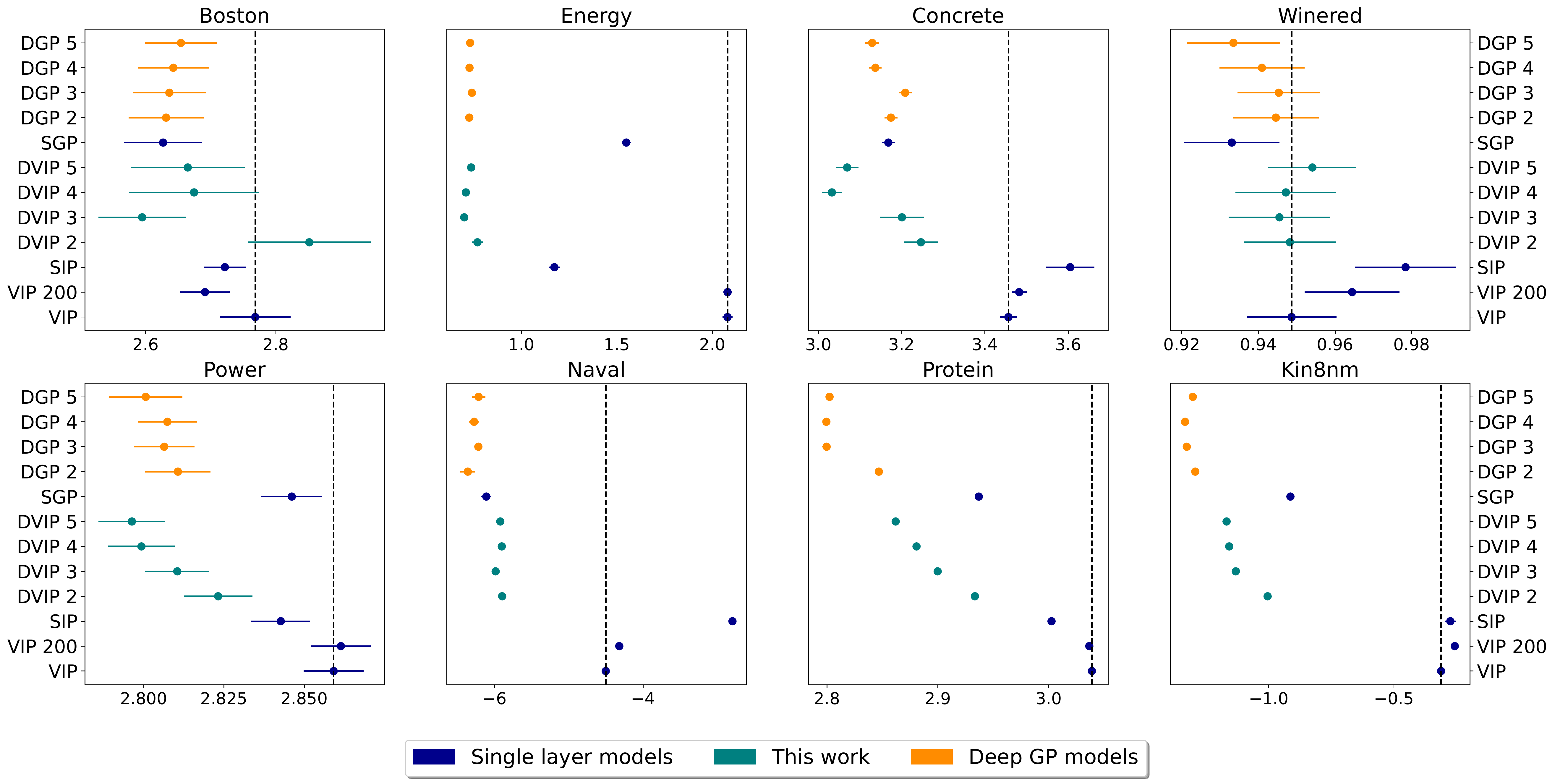}}
    \caption{Negative test log-likelihood results on \hyperref[para:uci]{regression UCI benchmark} datasets over 
	\(20\) splits. We show  standard errors. Lower values (to the left) are better.}
    \label{fig:uci_nll}
\end{figure}

\begin{figure}[htb!]
    \centering
    \makebox[\textwidth]{\includegraphics[width=0.9\textwidth]{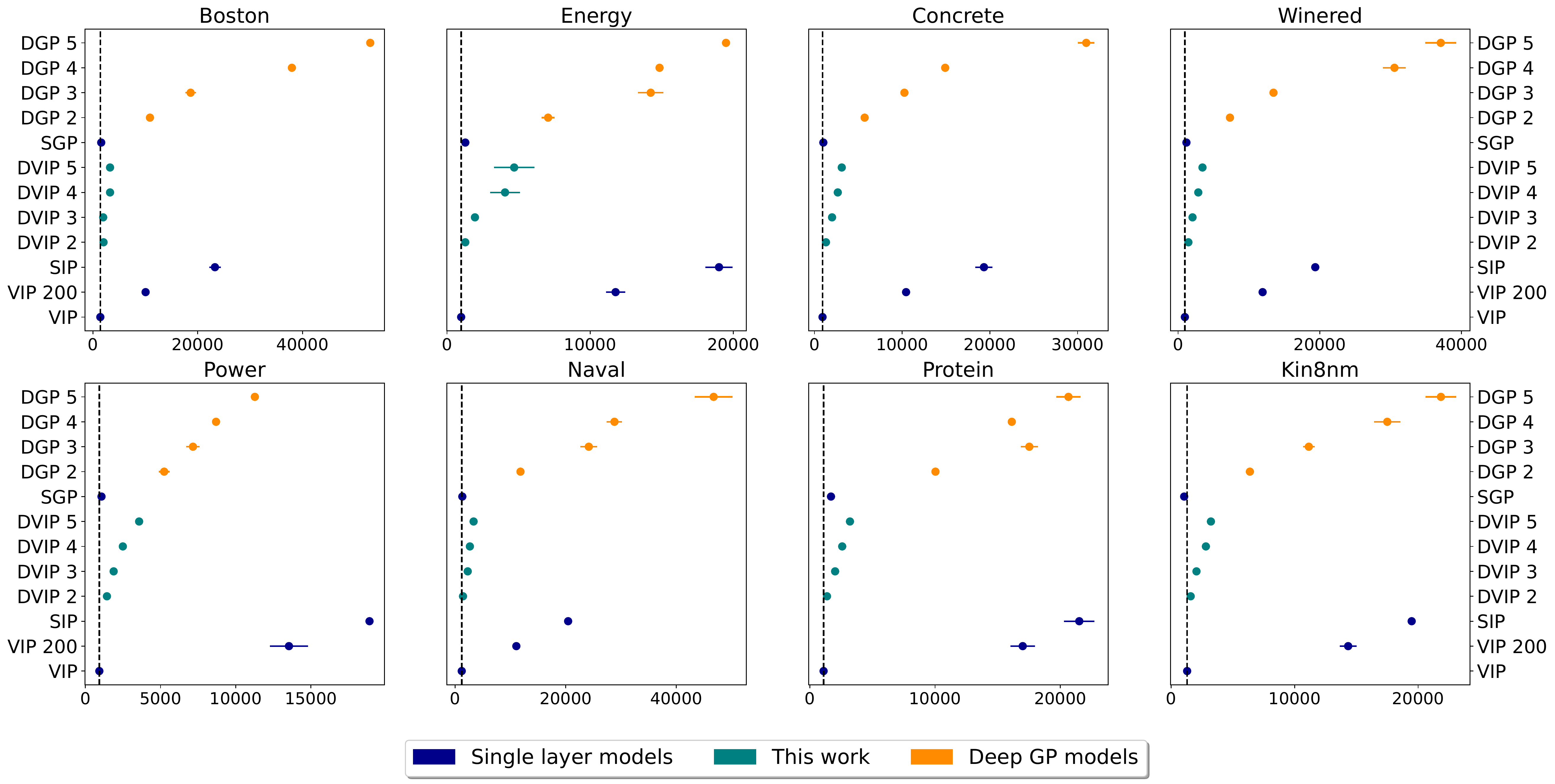}}
    \caption{CPU training time (in seconds) on \hyperref[para:uci]{regression UCI benchmark} datasets over \(20\) splits. We show standard errors. Lower values (to the left) are better.}
    \label{fig:uci_time}
\end{figure}

\paragraph{Regression UCI benchmarks.}\label{para:uci} We compare each method 
on 8 regression datasets from the UCI Repository \citep{uci}. 
Following common practice, we validate the performance using 20 different train / test 
splits of the data with \(10\%\) test size \citep{hernandez2015probabilistic}. 
We evaluate DVIP and DGP using 2, 3, 4 and 5 layers. We compare results with VIP, 
which is equivalent to DVIP with $L=1$, and with VIP using a bigger BNN 
of 200 units per layer. We also compare results with a single sparse GP, which 
is equivalent to DGP for $L=1$, and with a Sparse Implicit Processes (SIP) with the same prior \citep{santana2021sparse}. 
Figure \ref{fig:uci_nll} shows the results obtained in terms of the negative 
test log-likelihood. Results in terms of the RMSE and the exact figures are 
found in Appendix \ref{sec:app_extra_res}.
DVIP with at least 3 layers performs best on 4 out of the 8 datasets (\texttt{Boston}, \texttt{Energy}, 
\texttt{Concrete} and \texttt{Power}), having comparable results on \texttt{Winered} and 
\texttt{Naval} (all methods have zero RMSE on this dataset). 
DGPs perform best on 2 datasets (\texttt{Protein} and \texttt{Kin8nm}), but the 
differences are small. Appendix~\ref{sec:app_gp_prior} shows that using a GP prior 
in DVIP in these problems performs better at a higher cost. Adding more layers in DVIP 
does not lead to over-fitting and it gives similar and often better results 
(notably on larger datasets: \texttt{Naval}, \texttt{Protein}, \texttt{Power} and \texttt{Kin8nm}). 
DVIP also performs better than VIP and SIP most of the times. By contrast, using a more 
flexible BNN prior in VIP (\emph{i.e.}, 200 units) does not improve results.
Figure \ref{fig:uci_time} shows the training time in seconds of each method.
DVIP is faster than DGP and faster than VIP with the 200 units BNN prior. 
Summing up, DVIP achieves similar results to those of DGPs, but at a smaller cost.

\begin{figure}[htb!]
\centering
\begin{tabular}{cc}
  \includegraphics[width=.425\linewidth]{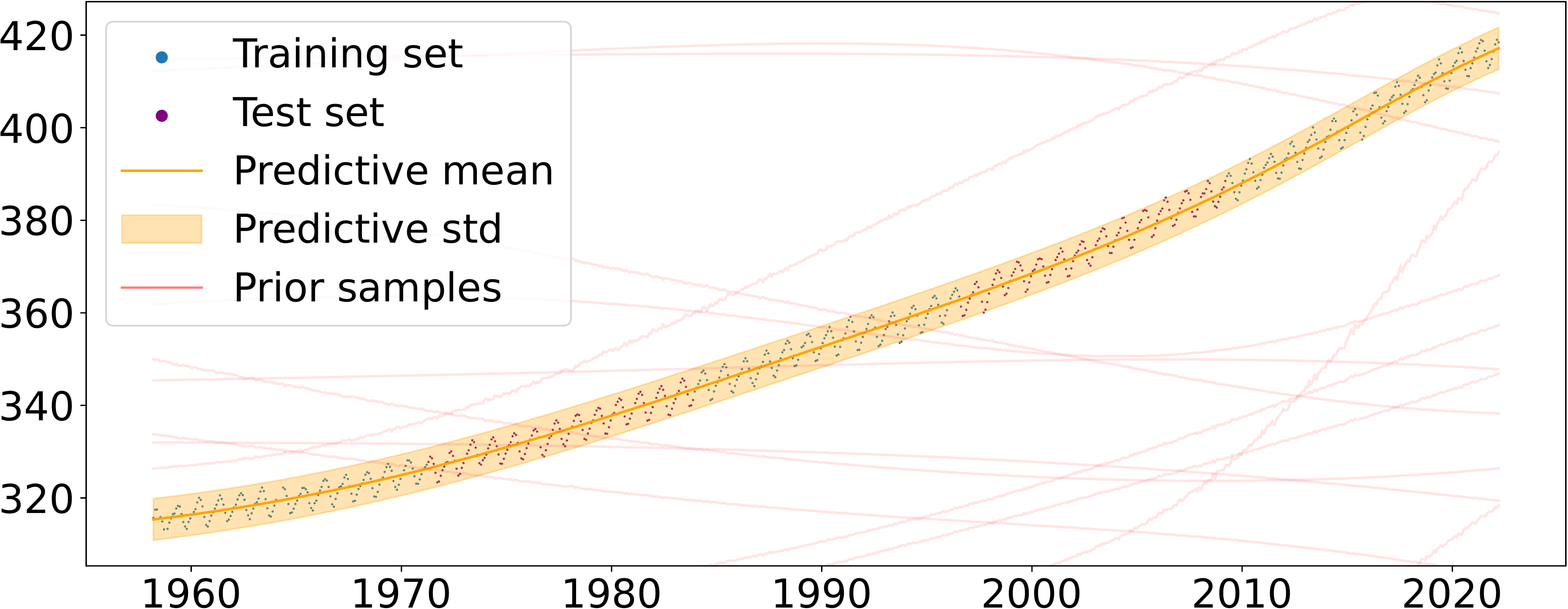} &
  \includegraphics[width=.425\linewidth]{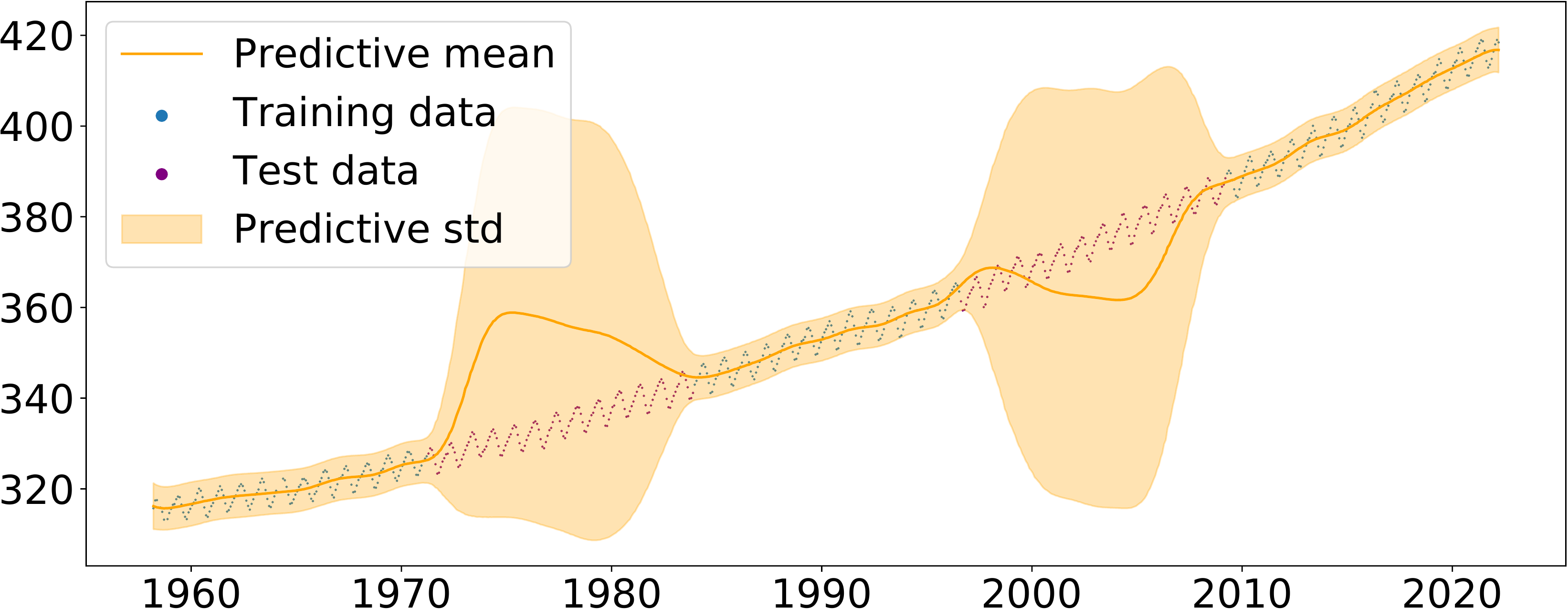}
\end{tabular}
\caption{\hyperref[para:co2]{Missing values interpolation} results on the \texttt{CO2} dataset. 
Predictive distribution of DVIP (left) and DGP (right) with 2 layers each. Two 
times the standard deviation is represented.}\label{fig:co2}
\end{figure}

\paragraph{Interpolation results.}\label{para:co2} We carry out experiments
on the CO2 time-series dataset (\url{https://scrippsco2.ucsd.edu}).
This dataset  has $\text{CO}_2$ measurements from the Mauna Loa Observatory, Hawaii, in 1978. 
We split the dataset in five consecutive and equal parts, and used the 2nd and 4th parts as 
test data. All models are trained for $100,000$ iterations.
Figure \ref{fig:co2} shows the predictive distribution of DVIP and DGP with $L=2$ on the data. 
DVIP captures the data trend in the missing gaps. For DVIP we show samples 
from the learned prior, which are very smooth. By contrast, a DGP with RBF kernels fails to 
capture the data trend, leading to mean reversion and over-estimation of the prediction 
uncertainty (similar results for SGP are shown in Appendix~\ref{sec:app_extra_res}). Thus, 
the BNN prior considered by DVIP could be a better choice here.  This issue of DGPs can 
be overcome using compositional kernels \citep{duvenaud2014avoiding}, but that requires using 
kernel search algorithms. 

\paragraph{Large scale regression.}\label{para:large_regression} We evaluate each method on 
3 large regression datasets. First, the \texttt{Year} dataset (UCI) with $515,345$
instances and \(90\) features, where the original train/test splits are used. Second, the US 
flight delay (\texttt{Airline}) dataset \citep{dutordoir2020sparse, hensman2017variational}, where 
following \cite{salimbeni2017doubly} we use the first $700,000$ instances for training and 
the next $100,000$ for testing. $8$ features are considered: \emph{month, day of month, 
day of week, plane age, air time, distance, arrival time and departure time}. For these two 
datasets, results are averaged over \(10\) different random seed initializations.
Lastly, we consider data recorded on January, 2015 from the \texttt{Taxi} 
dataset \citep{salimbeni2017doubly}. In this dataset $10$ attributes are considered: 
\emph{time of day, day of week, day of month, month, pickup latitude, pickup longitude, 
drop-off longitude, drop-off latitude, trip distance and trip duration}. Trips with a duration 
lower than 10 seconds and larger than 5 hours are removed as in \cite{salimbeni2017doubly}, 
leaving $12,695,289$ instances. Results are averaged over $20$ train/test 
splits with 90\% and 10\% of the data. Here, we trained each method for $500,000$ iterations.
The results obtained are shown in Table~\ref{tab:large_regression}. The last column shows the 
best result by DGP, which is achieved for $L=3$ on each dataset. We observe that 
DVIP outperforms VIP on all datasets, and on \texttt{Airline} and \texttt{Taxi}, the 
best method is DVIP. In \texttt{Taxi} a sparse GP and DGPs give similar results, 
while DVIP improves over VIP. The best method on \texttt{Year}, however, is 
DGP. The difference between DVIP and DGP is found in the prior (BNN vs. GP), and in the 
approximate inference algorithm (DGP uses inducing points for scalability and DVIP a linear model). 
Since DVIP generalizes DGP, DVIP using GP priors should give similar results to 
those of DGP on \texttt{Year}. Appendix~\ref{sec:app_gp_prior} shows, however, that the differences 
in \texttt{Year} are not only due to the chosen prior (BNN vs. GP) but also due to the 
posterior approximation (linear model vs. inducing points). VIP using inducing points and a GP prior 
gives similar results to those of SGP. The cost of approximately sampling from the GP prior 
in VIP is, however, too expensive to consider adding extra layers in DVIP. 

\begin{table}[htb!]
    \centering
    \scalebox{0.7}{
	\begin{tabular}{lr@{$\pm$}lr@{$\pm$}lr@{$\pm$}lr@{$\pm$}lr@{$\pm$}lr@{$\pm$}lr@{$\pm$}lr@{$\pm$}l}
        \toprule
        & \multicolumn{4}{c}{Single-layer} & \multicolumn{8}{c}{Ours} & \multicolumn{2}{c}{\citetalias{salimbeni2017doubly}}  \\ 
	\cmidrule(lr){2-5} \cmidrule(lr){6-13} \cmidrule(lr){14-15} 
         & \multicolumn{2}{c}{SGP} & \multicolumn{2}{c}{VIP} & \multicolumn{2}{c}{DVIP 2} & \multicolumn{2}{c}{DVIP 3} & 
		\multicolumn{2}{c}{DVIP 4} & \multicolumn{2}{c}{DVIP 5} & \multicolumn{2}{c}{DGP 3} \\
        \midrule
        Year &  9.15 & 0.01 & 10.27 & 0.01 & 9.61 & 0.03 & 9.34 & 0.02 & 9.30 & 0.03 & 9.27 & 0.03 & $\bm{8.94}$ & $\bm{0.03}$ \\
        Airline & 38.61 & 0.05 &  38.90 & 0.06 & 37.96 & 0.03 & 37.91 & 0.05 & 37.83 & 0.03 & $\bm{37.80}$ & $\bm{0.05}$ & 37.95&  0.04\\
        Taxi & 554.22&  0.32 & 554.60  & 0.19 & 549.28 & 0.59 & $\bm{531.42}$ &$\bm{1.59}$ & 547.33 &1.03 & 538.94 & 2.23 & 552.90 & 0.33\\
        \bottomrule
    \end{tabular}}
    \caption{Root mean squared error results on \hyperref[para:large_regression]{large scale 
	regression datasets.}}\label{tab:large_regression}
\end{table}

\paragraph{Image classification.}\label{para:image}
We consider the binary classification dataset 
\texttt{Rectangles} \citep{salimbeni2017doubly}
and the multi-class dataset \texttt{MNIST} \citep{deng2012mnist}. 
Each dataset has $28 \times 28$ pixels images.
The \texttt{Rectangles} dataset 
has $12,000$ images of a (non-square) rectangle. 
The task is to determine if the height is larger than the width. 
Here, we used a probit likelihood in each method.  The 
\texttt{MNIST} dataset has $70,000$ images of handwritten digits. 
The labels correspond with each digit. Here, we used the robust-max 
multi-class likelihood in each method \citep{hernandez2011robust}. In
\texttt{Rectangles}, $20,000$ iterations are enough to ensure convergence. We employ the provided train-test splits for each dataset. 
Critically, here we exploit DVIP's capability to use more flexible priors. 
In the first layer we employ a convolutional NN (CNN) prior with two layers of 4 and 8 
channels respectively. No input propagation is used in the first layer. The results 
obtained are shown in Table~\ref{tab:image}, averaged over \(10\) random seed initializations. 
We report the best obtained results for DGP, which are obtained for $L=3$.  
We observe that DVIP obtains much better results than those of DGP and VIP in terms of accuracy. 
DVIP increases accuracy by $11\%$ on \texttt{Rectangles} compared to DGP, probably as a consequence of the CNN prior considered 
in the first layer of the network being more suited for image-based datasets. Convolutional 
DGP can perform better than standard DGP in these tasks, however, the objective 
here is to highlight that DVIP allows to easily introduce 
domain-specific prior functions that might not be easily used by standard GPs. 
Image classification is only an example of this. 
Other examples may include using recurrent network architectures for sequential data.

\begin{table}[htb!]
    \centering
    \scalebox{0.65}{\begin{tabular}{lr@{$\pm$}lr@{$\pm$}lr@{$\pm$}lr@{$\pm$}lr@{$\pm$}lr@{$\pm$}lr@{$\pm$}lr@{$\pm$}lr@{$\pm$}l}
        \toprule
        \multirow{2.5}{*}{\textbf{MNIST}}& \multicolumn{4}{c}{Single-layer} & \multicolumn{4}{c}{Ours} & 
	\multicolumn{6}{c}{\citetalias{salimbeni2017doubly}}  \\ 
	\cmidrule(lr){2-5} \cmidrule(lr){6-9} \cmidrule(lr){10-15} 
	& \multicolumn{2}{c}{SGP} & \multicolumn{2}{c}{VIP}  
	& \multicolumn{2}{c}{DVIP 2} &  \multicolumn{2}{c}{DVIP 3}
        & \multicolumn{6}{c}{
		\begin{tabular}{c@{\hskip 0.4in}c}
		DGP 2 & DGP 3\\
		\end{tabular}
	} \\
        \midrule
        Accuracy (\%) & 96.25 & 0.04 & 97.99 & 0.03 & $\bm{98.39}$ & $\bm{0.05}$ &  98.36 & 0.04 & \multicolumn{6}{c}{
		\begin{tabular}{r@{$\pm$}l@{\hskip 0.4in}r@{$\pm$}l}
		97.75 & 0.04 &97.86 & 0.05 \\
		\end{tabular}
	} \\
        Likelihood & -0.146 & 0.00 & -0.144 & 0.00 & -0.074 & 0.00 & -0.080 & 0.00 & \multicolumn{6}{c}{
		\begin{tabular}{r@{$\pm$}l@{\hskip 0.25in}r@{$\pm$}l}
		-0.082 & 0.00 & $\bm{-0.072}$ & $\bm{0.00}$ \\
		\end{tabular}
	} \\
        \midrule \midrule
        \multirow{2.5}{*}{\textbf{Rectangles}}& \multicolumn{4}{c}{Single-layer} & \multicolumn{8}{c}{Ours} & \multicolumn{2}{c}{\citetalias{salimbeni2017doubly}}  \\ 
	\cmidrule(lr){2-5} \cmidrule(lr){6-13} \cmidrule(lr){14-15} 
	& \multicolumn{2}{c}{SGP} & \multicolumn{2}{c}{VIP}  
	& \multicolumn{2}{c}{DVIP 2} &  \multicolumn{2}{c}{DVIP 3}
        & \multicolumn{2}{c}{DVIP 4} & \multicolumn{2}{c}{DVIP 5} & \multicolumn{2}{c}{DGP 3}\\
        Accuracy (\%) & 72.54 & 0.14 & 85.63 & 0.18 &  87.84 & 0.20 & $\bm{88.21}$ & $\bm{0.12}$ & 87.43 &  0.20& 86.49 & 0.17 & 75.16 & 0.16\\
        Likelihood & -0.518 & 0.00 &  -0.348 & 0.00 & -0.306 & 0.00& $\bm{-0.295}$ &$\bm{0.00}$ &-0.309 & 0.00 & -0.320 & 0.00& -0.470 & 0.00\\
        AUC & 0.828 & 0.00 & 0.930 & 0.00 &0.950 & 0.00 & $\bm{0.953}$ & $\bm{0.00}$ & 0.947 & 0.00 & 0.939 &  0.00 & 0.858 &  0.00\\
        \bottomrule
    \end{tabular}
    }
    \caption{Results on \hyperref[para:image]{image classification datasets.}}\label{tab:image}
\end{table}

\paragraph{Large scale classification.}\label{para:large_class}
We evaluate each method on two massive binary datasets:  
\texttt{SUSY} and \texttt{HIGGS}, with \(5.5\) million and \(10\) million instances, respectively. 
These datasets contain Monte Carlo physics simulations to detect the presence of the 
Higgs boson and super-symmetry \citep{baldi2014searching}. We use the original 
train/test splits of the data, and train for $500,000$ iterations. We report the AUC 
metric for comparison with \cite{baldi2014searching,salimbeni2017doubly}. Results are 
shown in Table~\ref{tab:susy_higgs}, averaged over 10 different random seed initializations. 
In the case of DGPs, we report the best results, which correspond to $L=4$ and $L=5$, respectively.
We observe that DVIP achieves the highest performance on \texttt{SUSY} (AUC of \(0.8756\)) which 
is comparable to that of DGPs (\(0.8751\)) and to the best reported results in \cite{baldi2014searching}. Namely, 
shallow NNs (NN, \(0.875\)), deep NN (DNN, \(0.876\)) and boosted decision trees (BDT, \(0.863\)). 
On \texttt{HIGGS}, despite seeing an steady improvement over VIP by using additional 
layers, the performance is worse than that of DGP (AUC \(0.8324\)). Again, we believe 
that GPs with an RBF kernel may be a better prior here, and that 
DVIP using inducing points and a GP prior should give similar results to those 
of DGP. However, the high computational cost of approximately sampling from the GP 
prior will make this too expensive. 

\begin{table}[htb!]
    \centering
    \scalebox{0.65}{\begin{tabular}{lr@{$\pm$}lr@{$\pm$}lr@{$\pm$}lr@{$\pm$}lr@{$\pm$}lr@{$\pm$}lr@{$\pm$}lr@{$\pm$}lr@{$\pm$}l}
        \toprule
        \multirow{ 2.5}{*}{\textbf{SUSY}}& \multicolumn{4}{c}{Single-layer} & \multicolumn{8}{c}{Ours} & \multicolumn{2}{c}{\citetalias{salimbeni2017doubly}} \\ 
	\cmidrule(lr){2-5} \cmidrule(lr){6-13} \cmidrule(lr){14-15}
         & \multicolumn{2}{c}{SGP} & \multicolumn{2}{c}{VIP} & \multicolumn{2}{c}{DVIP 2} & 
		\multicolumn{2}{c}{DVIP 3} & \multicolumn{2}{c}{DVIP 4} & \multicolumn{2}{c}{DVIP 5} & \multicolumn{2}{c}{DGP 4}\\
        \midrule
        Accuracy (\%) & 79.75 & 0.02 & 78.68 & 0.02 &  80.11 & 0.03 & 80.13 & 0.01 & 80.22 & 0.01& $\bm{80.24}$ & $\bm{0.02}$ & 80.06 &  0.01\\
        Likelihood & -0.436 & 0.00 &  -0.456 & 0.00 & -0.429 & 0.00 & -0.429 & 0.00 & $\bm{-0.427}$ & $\bm{0.00}$ &  $\bm{-0.427}$ & $\bm{0.00}$& -0.432 & 0.00\\

        AUC &  0.8727 & 0.00  & 0.8572 & 0.00 & 0.8742 & 0.00 & 0.8749 & 0.00& 0.8755 & 0.00 & $\bm{0.8756}$ & $\bm{0.00}$ & 0.8751 & 0.00 \\
        \midrule\midrule
        \textbf{HIGGS} & \multicolumn{2}{c}{SGP} & \multicolumn{2}{c}{VIP} & \multicolumn{2}{c}{DVIP 2} & \multicolumn{2}{c}{DVIP 3} & \multicolumn{2}{c}{DVIP 4} & \multicolumn{2}{c}{DVIP 5} & \multicolumn{2}{c}{DGP 5}\\
        \midrule
        Accuracy (\%) & 69.95 & 0.03 & 57.42 & 0.03 & 66.09 & 0.02 & 69.85 & 0.02& 70.43 & 0.01& 72.01 & 0.02 & $\bm{74.92}$ & $\bm{0.01}$\\
        Likelihood & -0.573 & 0.00 &  -0.672 & 0.00 & -0.611 & 0.00& -0.575& 0.00 & -0.565 & 0.00 & -0.542 & 0.00 & $\bm{-0.501}$ & $\bm{0.00}$\\
        AUC &  0.7693 & 0.00 & 0.6247 & 0.00 & 0.7196 & 0.00 & 0.7704 & 0.00 &0.7782 & 0.00 & 0.7962 & 0.00 &  $\bm{0.8324}$ & $\bm{0.00}$\\
        \bottomrule
    \end{tabular}}
    \caption{Results on \hyperref[para:large_class]{large classification datasets.}}\label{tab:susy_higgs}
\end{table}

\paragraph{Impact of the Number of Samples $S$ and the Prior Architecture.} Appendix \ref{sec:app_samples} 
investigates the impact of the number of samples $S$ on DVIP's performance. 
The results show that one can get sometimes even better results in DVIP by increasing $S$ at the cost 
of larger training times. Appendix \ref{sec:app_archi} shows that changing the structure of 
the prior BNN does not heavily affect the results of DVIP.

%% file: sections/discussion.tex
Deep Variational Implicit Process (DVIP), a model based on the concatenation of implicit processes (IPs), is introduced as a flexible prior over latent functions. DVIP can be used on a variety of regression and classification problems with no need of hand-tuning. Our results show that DVIP outperforms or matches the performance of a single layer VIP and GPs. It also gives similar and sometimes better results than those of deep GPs (DGPs). However, DVIP has less computational cost when using a prior that is easy to sample from. Our experiments have also demonstrated that DVIP is both effective and scalable on a wide range of tasks. DVIP does not seem to over-fit on small datasets by increasing the depth, and on large datasets, extra layers often improve performance. We have also showed that increasing the number of layers is far more effective than increasing the complexity of the prior of a single-layer VIP model. Aside from the added computation time, which is rather minor, we see no drawbacks to the use of DVIP instead of a single-layer VIP, but rather significant benefits. 

The use of domain specific priors, such as CNNs in the first layer, has provided outstanding results in image-based datasets compared to other GP methods. This establishes a new use of IPs with not-so-general prior functions. We foresee employing these priors in other domain specific tasks, such as forecasting or data encoding, as an emerging field of study. The prior flexibility also results in a generalization of DGPs. As a matter of fact, DVIP gives similar results to those of DGPs if a GP is considered as the IP prior for each unit. Preliminary experiments in Appendix~\ref{sec:app_gp_prior} confirms this. 

Despite the good results, DVIP presents some limitations: first of all, the implicit prior works as a black-box from the interpretability point of view. The prior parameters do not represent a clear property of the model in comparison to kernel parameters in standard GPs. Furthermore, even though using $20$ samples from the prior has shown to give good results in some cases, there might be situations where this number must be increased, having a big impact in the model's training time. An unexpected result is that the cost of generating continuous samples from a GP prior in DVIP is too expensive. If a GP prior is to be used, it is cheaper to simply use a DGP as the underlying model.

%% file: appendixes/appendix.tex
\section{Derivation of the ELBO}
\label{sec:app_elbo}

The variational inference Evidence Lower BOund is defined as
\[
    \mathcal{L}\left(\Omega,\Theta, \{\bm \sigma_l^2\}_{l=1}^{L-1}\right) = \mathbb{E}_{q(\{\mathbf  F^l, \mathbf{A}^l\}_{l=1}^L)} \left[ \log \frac{p\left(\mathbf y, \{\mathbf  F^l, \mathbf{A}^l\}_{l=1}^L \right)}{q(\{\mathbf  F^l, \mathbf{A}^l\}_{l=1}^L)} \right]\,,
\]
where, using our model specification:
\begin{align*}
    p\left(\mathbf y, \{\mathbf  F^l, \mathbf{A}^l\}_{l=1}^L \right) &= 
	\prod_{n=1}^N p(y_n |\mathbf f_{\cdot,n}^L) \prod_{n=1}^N \prod_{l=1}^L 
	\prod_{h=1}^{H_l} p(f^l_{h,n} |\mathbf{a}^l_h)p(\mathbf a^l_h)\,,\\
	q(\{\mathbf  F^l, \mathbf{A}^l\}_{l=1}^L) & = 
	\prod_{n=1}^N \prod_{l=1}^L
        \prod_{h=1}^{H_l} p(f^l_{h,n} |\mathbf{a}^l_h)q(\mathbf a^l_h)\,.
\end{align*}
Using these expressions, the ELBO takes the following form:
\begin{align*}
    \mathcal{L} &=  \mathbb{E}_{q(\{\mathbf  F^l, \mathbf{A}^l\}_{l=1}^L)} \left[ \log \frac{p\left(\mathbf y, \{\mathbf  F^l, \mathbf{A}^l\}_{l=1}^L \right)}{q(\{\mathbf  F^l, \mathbf{A}^l\}_{l=1}^L)} \right]\\
    &=  \mathbb{E}_{q(\{\mathbf  F^l, \mathbf{A}^l\}_{l=1}^L)} \left[ \log \frac{\prod_{n=1}^N p(y_n |\mathbf f_{\cdot,n}^L) \prod_{n=1}^N \prod_{l=1}^L 
	\prod_{h=1}^{H_l} p(f^l_{h,n} |\mathbf{a}^l_h)p(\mathbf a^l_h)}{\prod_{n=1}^N \prod_{l=1}^L
        \prod_{h=1}^{H_l} p(f^l_{h,n} |\mathbf{a}^l_h)q(\mathbf a^l_h)} \right]\\
    &=  \mathbb{E}_{q(\{\mathbf  F^l, \mathbf{A}^l\}_{l=1}^L)} \left[ \log \frac{\prod_{n=1}^N p(y_n |\mathbf f_{\cdot,n}^L) \prod_{l=1}^L 
	\prod_{h=1}^{H_l}p(\mathbf a^l_h)}{\prod_{l=1}^L
        \prod_{h=1}^{H_l} q(\mathbf a^l_h)} \right]\,.
\end{align*}
The expectation can be split in two terms:
\[
    \mathcal{L} = \mathbb{E}_{q(\{\mathbf  F^l, \mathbf{A}^l\}_{l=1}^L)} \left[ \log \prod_{n=1}^N p(y_n |\mathbf f_{\cdot,n}^L) \right] 
    + 
    \mathbb{E}_{q(\{\mathbf  F^l, \mathbf{A}^l\}_{l=1}^L)} \left[ \log \frac{ \prod_{l=1}^L 
	\prod_{h=1}^{H_l}p(\mathbf a^l_h)}{\prod_{l=1}^L
        \prod_{h=1}^{H_l} q(\mathbf a^l_h)} \right]\,.
\]
The logarithm in the first term does not depend on the regression coefficients \(\{\bm A^l \}_{l=1}^L \) and neither on \(\{\bm F^l \}_{l=1}^{L-1}\). On the other hand, the logarithm on the second term does not depend on \(\{\bm F^l \}_{l=1}^{L}\)\,. Thus,
\begin{align*}
    \mathcal{L} &= \mathbb{E}_{q(\mathbf  F^L)} \left[ \log \prod_{n=1}^N p(y_n |\mathbf f_{\cdot,n}^L) \right] 
    + 
    \mathbb{E}_{q( \{\mathbf{A}^l\}_{l=1}^L)} \left[ \log \frac{ \prod_{l=1}^L 
	\prod_{h=1}^{H_l}p(\mathbf a^l_h)}{\prod_{l=1}^L
        \prod_{h=1}^{H_l} q(\mathbf a^l_h)} \right]\\
    &=\sum_{n=1}^N 
	\mathds{E}_q \big[ \log p\big(y_n|\mathbf{f}^L_{\cdot,n}\big)\big] -
	\sum_{l=1}^L \sum_{h=1}^{H_l} \text{KL}\big(q(\mathbf{a}_h^l) \big \rvert p(\mathbf{a}_h^l)\big)\,.
\end{align*}

\section{Derivation of the marginals}
\label{sec:app_marginals}

The variational distribution \(q(\{\bm f^l\}_{l=1}^L)\) factorizes as the product of Gaussian distributions:
\begin{align*}
    q(\{\mathbf  F^l\}_{l=1}^L) &=\prod_{n=1}^N \prod_{l=1}^L \prod_{h=1}^{H_L} q(f^l_{h,n}|\mathbf{f}^{l-1}_{\cdot,n})\\
    &= \prod_{n=1}^N \prod_{l=1}^L \prod_{h=1}^{H_L} \int p(f^l_{h,n}|\mathbf{f}^{l-1}_{\cdot,n}, \mathbf{a}^l_h) q(\mathbf{a}_h^l) d \mathbf{a}_h^l\\
    &= \prod_{n=1}^N \prod_{l=1}^L  \prod_{h=1}^{H_L}
	\mathcal{N}\left(f_{h,n}^l|\hat{m}_{h,n}^l(\mathbf{f}_{\cdot,n}^{l-1}), 
	\hat{v}_{h,n}^l(\mathbf{f}_{\cdot,n}^{l-1})\right)\,.
\end{align*}

As a result, the \(n^{th}\) marginal of the final layer depends only on the \(n^{th}\) marginals of the other layers. That is,
\[
    q( f_n^L) = \int \prod_{l=1}^L  \prod_{h=1}^{H_L}
	\mathcal{N}\left(f_{h,n}^l|\hat{m}_{h,n}^l(\mathbf{f}_{\cdot,n}^{l-1}), 
	\hat{v}_{h,n}^l(\mathbf{f}_{\cdot,n}^{l-1})\right) d \mathbf{f}_{\cdot,n}^1, \dots,d \mathbf{f}_{\cdot,n}^{L-1}\,.
\]

\section{Experimental Settings}
\label{sec:app_exp_settings}

To speed-up computations in DVIP, at each layer $l$ the generative function that 
defines the IP prior is shared across units. That is, the function 
$g_{\bm \theta_h^l}(\cdot, \bm z)$ is the same for every dimension 
$h$ in that layer. As a consequence, the prior IP samples only need to 
be generated once per layer, as in \cite{ma2019variational}. 
In the BNN prior of DVIP and VIP, we tune the prior mean and variance of each weight and bias 
by maximizing the corresponding estimate of the marginal likelihood. 
As no regularizer is used for the prior parameters, the prior mean and 
variances are constrained to be the same in a layer of the BNN. This 
configuration avoids over-fitting and leads to improved results. The positive effect 
of this constraint is shown in Appendix~\ref{sec:app_prior_constraint}. 
In DGP we consider $100$ shared inducing points 
in each layer. We use ADAM \citep{kingma2015} as the optimization algorithm, and we set the learning 
rate to $10^{-3}$, in DVIP, as in \cite{ma2019variational}. In DGP we use $10^{-2}$ as the learning 
rate, as in \cite{salimbeni2017doubly}. Unless indicated otherwise, in DVIP 
and DGP we use the input dimension as the layer dimensionality, \emph{i.e} $H_l = D$, for $l=1,\dots,L-1$. 
In DGP the kernel employed is RBF with ARD \citep{rasmussen2005book}.
The batch size is $100$. All methods are trained for $150,000$ iterations unless indicated otherwise.
In VIP we do not employ the marginal likelihood regularizer described in \cite{ma2019variational}, 
since the authors of that paper told us that they did not use it in practice.
Similarly, we do not regularize the estimation of the prior IP covariances in VIP nor DVIP.
The inner dimensions of DVIP and DGP are fixed to the minimum between the number of attributes of the dataset 
and $30$, as in \cite{salimbeni2017doubly}. We use $\alpha = 0.5$ for VIP, as suggested in \cite{ma2019variational}. The reason 
why \(\alpha = 0\) is used in the experiments on DVIP is that the use of alpha-divergences 
requires the optimization of the hyper-parameter \(\alpha\), which was against 
our \emph{no hand-tuning} approach. Moreover, even if a fixed value of \(\alpha\) could 
perform well on average, its use would also require to propagate more than one Monte 
Carlo sample to get a low biased estimate of objective function, which slows down training. 
For these reasons, we decided to keep the standard VI ELBO. Future work may consider using alpha-divergences for 
training DVIP. The source code can be accessed using the Github Repository \emph{\href{https://github.com/Ludvins/DeepVariationalImplicitProcesses}{DeepVariationalImplicitProcesses}}.

\section{Impact of the Constrained Prior}\label{sec:app_prior_constraint}

\begin{figure}[htp!]
    \centering
      \includegraphics[width=.7\linewidth]{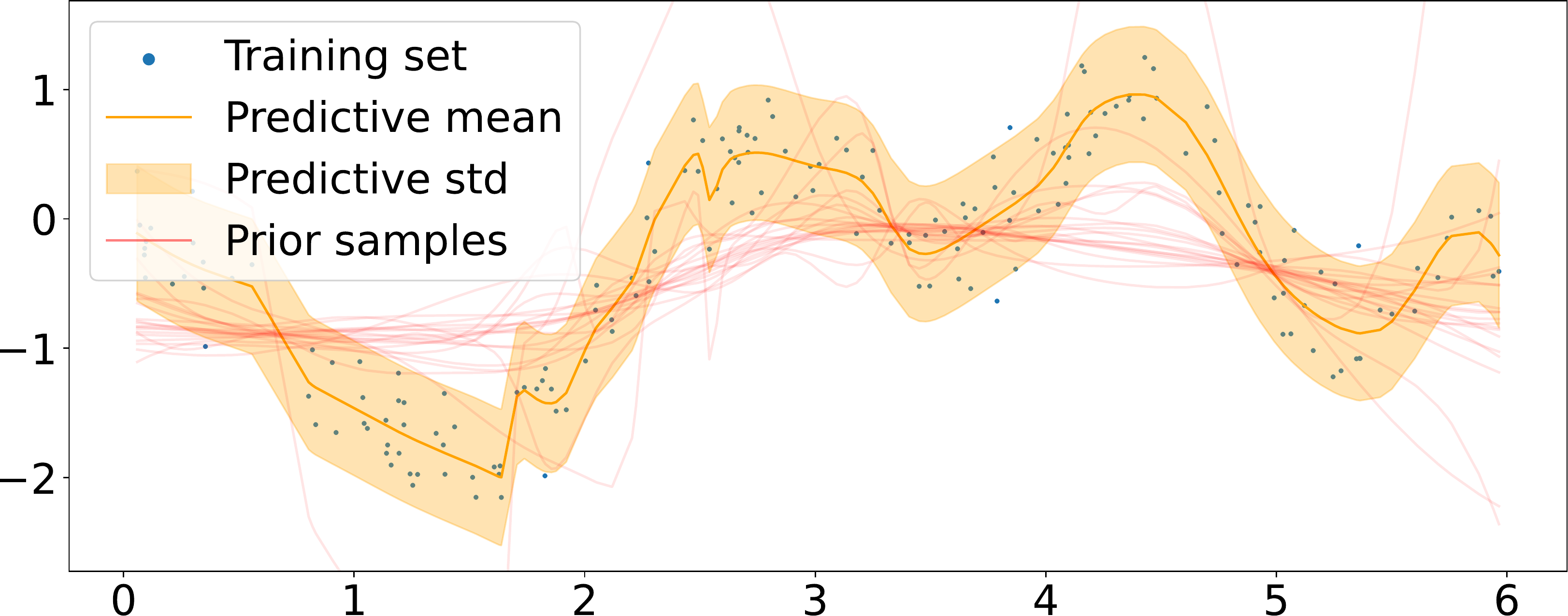}
      \includegraphics[width=.7\linewidth]{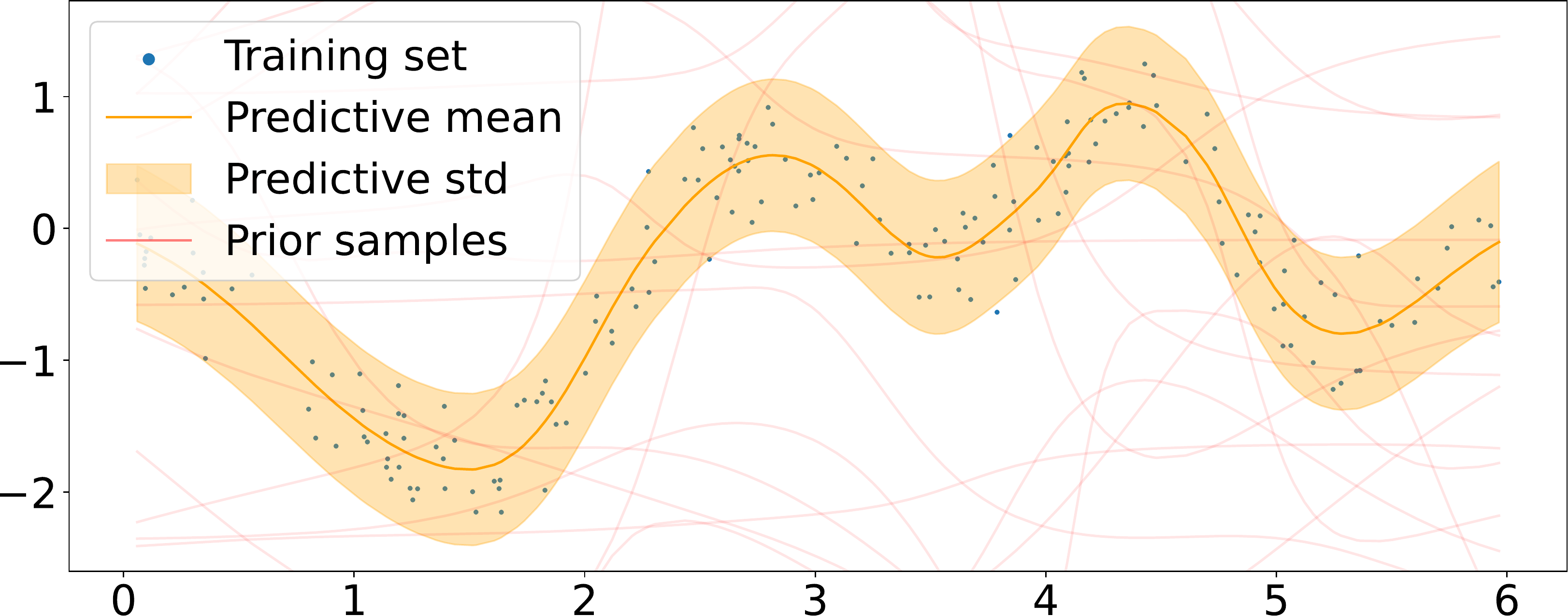}
    \caption{Resulting predictive distribution and prior samples over a toy dataset with full BNN prior (above) and constraint prior BNN (below).}\label{fig:vip_prior}
\end{figure}

Figure~\ref{fig:vip_prior} shows, on a toy problem, the obtained predictive distributions and learned prior 
samples of VIP, for $\alpha = 0$. This corresponds to a 1 layer particular case of DVIP.
We consider two cases: (1) using a full unconstrained prior BNN, in which the mean and variance of each
weight and bias can be independently tuned, (shown above) and (2), the considered constrained 
BNN in which prior means and variances are shared across layers 
(shown below). The second approach considerably reduces the number of parameters in 
the prior and we observe that  it generates smoother prior functions. The predictive distribution 
is also smoother than when the prior is unconstrained. However, despite providing better results by avoiding 
over-fitting, there might be datasets where using the full unconstrained parameterization of the BNN leads to 
improved results. For example, in problems where a more flexible prior may be beneficial to obtain good 
generalization properties on un-seen data.

Consider the Boston and the Energy datasets. 
To highlight that prior over-fitting is a dataset-specific matter, 
Table \ref{tab:prior_overfitting} shows the obtained results using an unconstrained prior 
and a constrained prior for VIP on the aforementioned datasets. As one may see, significantly over-fitting is taking place 
on Boston. More precisely, the training error improves a lot when using the un-constrained prior. By contrast, test
error and other performance metrics deteriorate on the test set. In Energy, however, a more flexible prior 
results in better test RMSE and CRPS, but worse test log-likelihood. 

\begin{table}[H]
    \centering
    \scalebox{0.8}{\begin{tabular}{lcccccc}
        \toprule
         \textbf{Unconstrained }& RMSE Train & RMSE Test & NLL Train & NLL Test & CRPS Train & CRPS Test \\
        \midrule
        Boston & \(0.05 \pm 0.00\) & \(5.85 \pm 0.14\) & \(-1.21 \pm 0.19\) & \(5126.07 \pm 274.79\) &\(0.03 \pm 0.00\) & \(4.31 \pm 0.08\) \\
        Energy & \(0.14 \pm 0.00\) & \(0.57 \pm 0.01\) & \(-0.51 \pm 0.01\) & \(6.52 \pm 0.42\) & \(	0.079 \pm 0.00\) & \(	0.36 \pm 0.01\)\\
        \midrule\midrule
         \textbf{Constrained }& RMSE Train & RMSE Test & NLL Train & NLL Test & CRPS Train & CRPS Test \\
        \midrule
        Boston & \(3.90 \pm 0.02\) & \(4.73 \pm 0.24\) & \(2.77 \pm 0.00\) & \(23.03 \pm 0.07\) &\(2.06 \pm 0.01\) & \(2.40 \pm 0.08\) \\
        Energy & \(2.35 \pm 0.03\) & \(2.57 \pm 0.08\) & \(2.27 \pm 0.01\) & \(2.07 \pm 0.02\) & \(1.28 \pm 0.01\) & \(1.27 \pm 0.04\)\\
        \bottomrule
    \end{tabular}}
    \caption{Results on Boston and Energy dataset using the constrained and unconstrained BNN prior for VIP with \(\alpha = 0\).}\label{tab:prior_overfitting}
\end{table}

\section{Impact of the number of prior samples}
\label{sec:app_samples}

We explore the impact of the number of prior samples \(S\) has on the performance and training time of the proposed method. Table~\ref{tab:prior_samples} shows the results obtained using DVIP with 3 layers on Protein and Power datasets (UCI). These results show how increasing the number of prior samples produces better results, in terms of performance, compared to using lower values of \(S\). However, it is important to consider that this value scales quadratically the computational complexity of evaluating the ELBO, heavily influencing the training cost of the model.

\begin{table}[H]
    \centering
    \scalebox{0.85}{\begin{tabular}{l*{9}{c}}
        \toprule
        \textbf{Power} & \(S = 10\) & \(S = 20\)& \(S = 30\)& \(S = 40\)& \(S = 50\) \\
        \midrule
        RMSE & \(4.03 \pm 0.04\) & \(4.01 \pm 0.04\) & \(3.94 \pm 	0.04\) & \(3.92 \pm 0.04\) & \( 	3.94 \pm 0.04\)\\
        NLL & \(2.81 \pm 0.01\) & \(2.81 \pm 0.00\) & \(2.79 \pm 0.01\) & \(2.78 \pm 0.01\) & \(2.79 \pm	0.01\)\\
        CRPS & \(2.19 \pm 0.01\) & \(2.18 \pm 0.01\) & \(2.14 \pm 0.01\) & \(2.11 \pm	0.01\) & \(2.13 \pm 0.01\)\\
        CPU Time (s) & \(2693 \pm 19\) & \(2806 \pm 22\) & \(3152 \pm 20\) & \(3451 \pm 63\) & \(3742 \pm 55\)\\ 
        \midrule
        \midrule
        \textbf{Protein} & \(S = 10\) & \(S = 20\)& \(S = 30\)& \(S = 40\)& \(S = 50\) \\
        \midrule
        RMSE & \(4.53 \pm 0.01\) & \(4.40 \pm 0.01\) & \(4.28 \pm 0.01\) & \(4.27 \pm 0.01\) & \( 	4.21 \pm 0.01\)\\
        NLL & \(2.92 \pm 0.00\) & \(2.90 \pm 0.00\) & \(2.87 \pm 0.00\) & \(2.86 \pm 0.00\) & \(2.85 \pm 0.00\)\\
        CRPS & \(2.52 \pm 0.00\) & \(2.43 \pm 0.00\) & \(2.36 \pm 0.00\) & \(2.34 \pm 0.00\) & \(2.31 \pm 0.00\)\\
        CPU Time (s) & \(2334 \pm 28\) & \(2734 \pm 19\) & \(3616 \pm 11\) & \(3727 \pm 35\) & \(4330 \pm 52\)\\ 
        \bottomrule
    \end{tabular}}
    \caption{Results on Power and Protein datasets (UCI) using DVIP with 3 layers and different values of the number of prior samples \(S\).}\label{tab:prior_samples}
\end{table}

\section{Robustness over the Prior BNN Architecture}
\label{sec:app_archi}

In this section, we study the impact that changing the prior BNN structure has over the performance of the proposed method.  Table~\ref{tab:prior_structure} shows the results obtained using DVIP with 2,3 and 4 layers on Protein and Power datasets (UCI) with two different BNN structures, 2 hidden layers with 20 units (20-20) and 3 hidden layers with 10 units (10-10-10). These results (that are to be compared with the original ones obtained using 2 hidden layers and 10 units on Table~\ref{tab:full_uci}) show that changing the structure of the BNN does not heavily affect the obtained results, given that it is capable of learning similar function distributions with the different BNN architectures.

\begin{table}[H]
    \centering
    \scalebox{0.85}{\begin{tabular}{lcccccc}
        \toprule
        \textbf{} & \multicolumn{3}{c}{BNN 10-10-10} & \multicolumn{3}{c}{BNN 20-20}  \\
        \midrule
        \textbf{Power} & DVIP \(2\) & DVIP \(3\) & DVIP \(4\)& DVIP \(2\) & DVIP \(3\) & DVIP \(4\)\\
        \midrule
        RMSE & \(4.02 \pm 0.04\) & \(3.94 \pm 0.04\) & \(3.97 \pm 0.04\) & \(4.02 \pm 0.04\) & \( 3.99 \pm 0.04\)& \( 3.93 \pm 0.04\)\\
        NLL & \(2.81 \pm 0.01\) & \(2.79 \pm 0.00\) & \(2.80 \pm 0.01\) & \(2.81 \pm 0.01\) & \(2.80 \pm	0.01\)& \(2.79 \pm	0.01\)\\
        CRPS & \(2.18 \pm 0.01\) & \(2.13 \pm 0.01\) & \(2.15 \pm 0.01\) & \(2.18 \pm 0.01\) & \(2.16 \pm 0.01\) & \(2.13 \pm 0.01\) \\
        \midrule
        \midrule
        \textbf{Protein} & DVIP \(2\) & DVIP \(3\) & DVIP \(4\)& DVIP \(2\) & DVIP \(3\) & DVIP \(4\)\\        \midrule
        RMSE & \(4.57 \pm 0.01\) & \(4.37 \pm 0.01\) & \(4.29 \pm 0.01\) & \(4.55 \pm 0.01\) & \(4.43 \pm 0.01\)& \(4.31 \pm 0.01\)\\
        NLL & \(2.93\pm 0.00\) & \(2.89 \pm 0.00\) & \(2.87 \pm 0.00\) & \(2.93 \pm 0.00\) & \(2.90 \pm 0.00\) & \(2.87 \pm 0.00\)\\
        CRPS & \(2.55 \pm 0.00\) & \(2.42 \pm 0.00\) & \(2.36 \pm 0.00\) & \(2.54 \pm 0.00\) & \(2.45 \pm 0.00\) & \(2.37 \pm 0.00\)\\
        \bottomrule
    \end{tabular}}
    \caption{Results on Power and Protein datasets (UCI) using DVIP with different prior BNN architectures.}\label{tab:prior_structure}
\end{table}

\section{Using a GP as prior in DVIP and VIP}\label{sec:app_gp_prior}

In this section we investigate the use of a GP prior in DVIP to approximate GPs 
and hence DGPs. From a theoretical point of view, GP samples could 
be used in VIPs prior, ensuring that VIP does converge to a GP when the number 
of prior samples (and linear regression coefficients) is large enough. As a result, 
DVIP can converge to a DGP. However, DVIP needs to evaluate covariances among the
process values at the training points. This requires taking continuous samples from a GP,
something that is not possible in practice unless one samples the process values at all the 
training points, something that is intractable for big datasets. To surpass this 
limitation, a BNN with a single layer of cosine activation functions can be used to approximate 
a GP with RBF kernel \citep{rahimi2007}. Generating continuous samples from this BNN is easy. One only has 
to sample the weights from their corresponding prior distribution. However, in order to correctly approximate 
the GP, a large number of hidden units is needed in the BNN, increasing the computational cost 
of taking such samples. 

Furthermore, in many situations the predictive distribution of a sparse GP is very 
different of that of a full GP. Meaning that even when using an approximate
GP prior in VIP, by means of a BNN with enough hidden units, it may not be enough to accurately 
approximate a sparse GP. Specifically, we have observed that this is the case in the Year dataset. 
In this dataset, the difference in performance between DVIP and DGP is not only a consequence of the 
different prior, but also the posterior approximation. More precisely, DVIP uses a linear regression 
approximation to the GP, while a DGP uses an inducing points based approximation. To show this, 
we have also implemented an inducing points approximation on VIP. For this, the required covariance 
matrices are estimated using a large number of prior samples. The obtained results can be seen in 
Table~\ref{tab:gp_prior}. There, we show the results of VIP using an approximated GP prior, using both the 
linear regression approximation and an approximation based on inducing points. We also report the results of the 
sparse GP and the average training time of each method in seconds.
The table shows that VIP can be used with inducing points to approximate a GP. 
Specifically, the inducing points approximation of VIP gives very similar results to those of a sparse GP.
However, this is achieved at at a prohibitive training time. The 
computational bottlenecks are the GP approximation using a wide single layer BNN (we used 2000 units in the 
hidden layer), and the 
generation of a large number of prior samples from the BNN to approximate the covariance matrix 
(we used 2000 samples). Given the high computational cost of training a VIP model on this dataset, considering
DVIP with more than 1 layer is too expensive. 

\begin{table}[H]
    \centering
    \scalebox{0.8}{\begin{tabular}{l*{8}{c}}
        \toprule
        \textbf{Year} & VIP & VIP-GP (linear regression) & VIP-GP (inducing points) & SGP \\
        \midrule
        RMSE & \(10.27 \pm 0.01\) & \(10.23 \pm 0.01\) & \(9.28 \pm 0.01\) & \(9.15 \pm 0.01\) \\
        NLL & \(3.74 \pm 0.00\) & \(3.77 \pm 0.00\) & \(3.64 \pm 0.00\) & \(3.62 \pm 0.00\)\\
        CRPS & \(5.45 \pm 0.01\) & \(5.45 \pm 0.02\) & \(4.85 \pm 0.01 \) & \(4.83 \pm 0.01\)\\
        CPU Time (s) & \(1217 \pm 257\) & \(1687\pm 271\) & \(30867 \pm 326\) & \(1874 \pm 265\)\\ 
        \bottomrule
    \end{tabular}}
    \caption{Results on Year dataset of VIP with the usual BNN prior with \(2\) hidden units of 
	width \(10\) and tanh activations, VIP using a BNN that approximates a GP 
	with RBF kernel, VIP with the last prior and \(100\) inducing points and a 
	sparse GP with \(100\) inducing points. Experiments with VIP are trained using \(\alpha = 0\). }\label{tab:gp_prior}
\end{table}

We have carried out extra experiments in the smaller datasets Protein and Kin8nm to assess if using
a GP prior in the context of DVIP generates results that are closer to those of a DGP.
These are the regression datasets from the UCI repository where DGPs performed better than DVIP.
In this case, we did not consider the inducing points approximation of the GP, as in the previous paragraph,
but the linear regression approximation of VIP. We include results for (i) DVIP using the initially proposed BNN prior 
with 2 layers of 10 hidden units and tanh activation functions; (ii) DVIP using the wide BNN that approximates
the prior GP; (iii) DGPs using sparse GPs in each layer based on inducing points. 
The results obtained are shown in Tables \ref{tab:Protein} and \ref{tab:Kin8nm}. 
We observe that in both cases using a GP prior in DVIP often improves results and performs similarly to a DGP,
especially for a large number of layers $L$, even when there are differences in the approximation of 
the posterior distribution, \emph{i.e.}, DVIP uses a linear regression approximation to the GP and 
DGP uses inducing points. Again, using a wide BNN to approximate the prior GP results in a
significant increment of the training time, making DVIP slower than DGP.

\begin{table}[H]
    \centering
    \scalebox{0.8}{\begin{tabular}{l*{9}{c}}
        \toprule
        \textbf{BNN Prior} & VIP & DVIP 2 & DVIP 3 & DVIP 4 & DVIP 5 \\
        \midrule
        RMSE & \(4.76 \pm 0.01\) & \(4.24 \pm  0.01\)&  \(	4.14 \pm  0.01\)&  \(4.14 \pm  0.01\)&  \(\bm{4.09 \pm  0.01}\)&  \\
        NLL & \(2.98 \pm 0.00\)& \(2.86 \pm 0.00\)&  \(2.84 \pm  0.00\)&  \(2.84 \pm  0.00\)&  \(\bm{2.83 \pm  0.00}\)&  \\
        CRPS & \(2.68 \pm 0.00\) &\(2.34 \pm  0.00\)&  \(	2.26 \pm  0.00\)&  \(	2.25 \pm  0.00\)&  \(\bm{2.21 \pm  0.00}\) & \\
        CPU time (s) & \(3086 \pm  173\) & \(3981 \pm  182\)& \(8604 \pm  774\)& \(9931 \pm  616\)& \(	12568 \pm  327\)\\
        \midrule
        \textbf{GP Prior} & VIP & DVIP 2 & DVIP 3 & DVIP 4 & DVIP 5 \\
        \midrule
        RMSE & \(4.89\pm  0.01\) & \(4.26\pm  0.01\)&  \(4.07\pm  0.01\) & \(4.02\pm  0.01\) & \(\bm{4.01\pm  0.01}\)\\
        NLL & \(3.00 \pm  0.00\)& \(2.87 \pm 0.00\)& \(2.82 \pm 0.00\)& \(2.81 \pm 0.00\)& \(\bm{2.81 \pm 0.00}\)    \\
        CRPS & \(2.77 \pm  0.00\) &\(2.35 \pm  0.00\)& \(2.21 \pm  0.00\)&\(2.17 \pm  0.00\)&\(\bm{2.16 \pm  0.00}\)\\
        CPU time (s) & \(5880\pm  249\) & \(12293\pm  564\) & \(20351\pm  1110\) & \(18514\pm  575\) & \(	25835\pm  1259\)\\
        \midrule
        \citetalias{salimbeni2017doubly} & SGP & DGP 2 & DGP 3 & DGP 4 & DGP 5 \\
        \midrule
        RMSE & \(4.56 \pm 0.01\) & \( 4.17 \pm  0.01\) &  \(\bm{4.00 \pm  0.01}\)&  \( 4.01 \pm  0.01\)&  \(4.02 \pm  0.01\)\\
        NLL & \(2.93\pm 0.00\) & \( 2.84 \pm  0.00\)&  \(\bm{2.79 \pm  0.00}\)&  \(\bm{2.79 \pm  0.00}\)&  \(2.80 \pm  0.00\) \\
        CRPS & \(2.56 \pm 0.00\) & \( 2.31 \pm  0.00\)&  \(\bm{2.19 \pm  0.00}\)&  \(\bm{2.19 \pm  0.00}\)&  \(2.20 \pm  0.00\) \\
        CPU time (s) & \(2690 \pm 114\) & \(10031 \pm 129\) & \(17528\pm 689\) & \(16128\pm 190\) & \(20653\pm 969\)\\
        \bottomrule
    \end{tabular}}
    \caption{Results on Protein UCI dataset using DVIP with the usual BNN prior, the approximated GP prior and deep GPs. Experiments with VIP are trained using \(\alpha = 0\). }\label{tab:Protein}
\end{table}

\begin{table}[H]
    \centering
    \scalebox{0.8}{\begin{tabular}{l*{9}{c}}
        \toprule
        \textbf{BNN Prior} & VIP & DVIP 2 & DVIP 3 & DVIP 4 & DVIP 5 \\
        \midrule
        RMSE & \(0.15 \pm 0.00\) & \(\bm{0.07 \pm  0.00}\)&  \(	\bm{0.07 \pm  0.00}\)&  \(\bm{0.07 \pm  0.00}\)&  \(\bm{0.07 \pm  0.00}\)&  \\
        NLL & \(-0.47 \pm 0.00\)& \(-1.13 \pm 0.00\)&  \(\bm{-1.18 \pm  0.00}\)&  \(-1.16 \pm  0.00\)&  \(-1.17 \pm  0.00\)&  \\
        CRPS & \(0.08 \pm 0.00\) &\(\bm{0.04 \pm  0.00}\)&  \(\bm{0.04 \pm  0.00}\)&  \(	\bm{0.04 \pm  0.00}\)&  \(\bm{0.04 \pm  0.00}\) & \\
        CPU time (s) & \(2109 \pm 57\) & \(5086 \pm 232\)& \(6927 \pm 27\)& \(9459 \pm 77\)& \(11763 \pm 141\)\\
        \midrule
        \textbf{GP Prior} & VIP & DVIP 2 & DVIP 3 & DVIP 4 & DVIP 5 \\
        \midrule
        RMSE & \(0.14 \pm 0.00\) & \(0.07 \pm  0.00\)&  \(\bm{0.06 \pm  0.00}\)&  \(\bm{0.06 \pm  0.00}\)&  \(\bm{0.06 \pm  0.00}\) \\
        NLL & \(-0.43\pm 0.00\) & \(-1.20 \pm  0.00\)&  \(-1.25 \pm  0.00\)&  \(\bm{-1.26 \pm  0.00}\)&  \(- 1.25 \pm  0.00\) \\
        CRPS & \(0.08 \pm 0.00\) &\(0.04 \pm  0.00\)&  \(\bm{0.03 \pm  0.00}\)&  \(\bm{0.03 \pm  0.00}\)&  \(\bm{0.03 \pm  0.00}\) \\
        CPU time (s) & \(6672 \pm  442\) & \(14300 \pm  847\) & \(17573 \pm  1033\) & \(22561 \pm  980\) & \(	22669 \pm  657\)\\
        \midrule
        \citetalias{salimbeni2017doubly} & SGP & DGP 2 & DGP 3 & DGP 4 & DGP 5 \\
        \midrule
        RMSE & \(0.09 \pm 0.00\) & \(\bm{0.06 \pm  0.00}\)&  \(\bm{0.06 \pm  0.00}\)&  \(\bm{0.06 \pm  0.00}\)&  \(\bm{0.06 \pm  0.00}\) \\
        NLL & \(-0.91\pm 0.00\) & \(-1.29 \pm  0.00\)&  \(-1.32 \pm  0.00\)&  \(\bm{-1.33 \pm  0.00}\)&  \(- 1.30 \pm  0.00\) \\
        CRPS & \(0.05 \pm 0.00\) &\(	\bm{0.03 \pm  0.00}\)&  \(\bm{0.03 \pm  0.00}\)&  \(\bm{0.03 \pm  0.00}\)&  \(\bm{0.03 \pm  0.00}\) \\
        CPU time (s) & \(2053 \pm  81\) & \(6375 \pm  331\) & \(11147 \pm  472\) & \(17502 \pm  1060\) & \(21846 \pm  1246\)\\
        \bottomrule
    \end{tabular}}
    \caption{Results on Kin8nm UCI dataset using DVIP with the usual BNN prior, the approximated GP prior and deep GPs. Experiments with VIP are trained using \(\alpha = 0\). }\label{tab:Kin8nm}
\end{table}

As a conclusion, DVIPs general and flexible definition allows the usage of (approximate) 
GP priors and inducing points. This enables performing (nearly) equally to a sparse GP or a DGP. 
However, the computational overhead of doing these approximations is prohibitive. Thus, 
if a GP prior is to be considered, it is a much better approach to just use the DGP framework, 
and leave DVIP to cases where flexible but easy-to-sample-from priors can be used.

\section{Further results}
\label{sec:app_extra_res}

The results regarding the \hyperref[para:uci]{regression UCI benchmark datasets} are provided in Table~\ref{tab:full_uci}. In addition, results on \hyperref[para:large_regression]{large scale regression datasets} are given in Table~\ref{tab:full_large_regression}.

\begin{table}[H]
    \centering
    \scalebox{0.7}{\begin{tabular}{l*{9}{c}}
        \toprule
        \multirow{ 2.5}{*}{\textbf{NLL}}& \multicolumn{2}{c}{Single-layer} & \multicolumn{4}{c}{Ours} & \citetalias{salimbeni2017doubly}  \\ \cmidrule(lr){2-3} \cmidrule(lr){4-7} \cmidrule(lr){8-8}
         & SGP & VIP & DVIP 2 & DVIP 3 & DVIP 4 & DVIP 5 & Best DGP \\
        \midrule
        Year & \(3.62 \pm 0.00 \) & \(3.74 \pm 0.00\) & \(3.68 \pm 0.00\) & \(3.64 \pm 0.00\) & \(3.64 \pm 0.00\) & \(3.63 \pm 0.00\) & \(\bm{3.59 \pm 0.00}\)\\
        Airline & \(5.10 \pm 0.00\) &  \(5.11 \pm 0.00\) & \(5.08 \pm 0.00\)& \(5.07 \pm 0.00\) & \(5.07 \pm 0.00\)& \(\bm{5.06 \pm 0.00}\)& \(5.07 \pm 0.00\)\\
        Taxi & \(7.73 \pm 0.00\) & \(7.73 \pm 0.00\) & \(7.72 \pm 0.00\) & \(\bm{7.69 \pm 0.00}\) & \(7.72 \pm 0.00\) & \(7.70 \pm 0.00\) & \(7.73 \pm 0.00\)\\
        \midrule \midrule
        \multirow{ 2.5}{*}{\textbf{CRPS}}& \multicolumn{2}{c}{Single-layer} & \multicolumn{4}{c}{Ours} & \citetalias{salimbeni2017doubly}  \\ \cmidrule(lr){2-3} \cmidrule(lr){4-7} \cmidrule(lr){8-8}
         & SGP & VIP & DVIP 2 & DVIP 3 & DVIP 4 & DVIP 5 & Best DGP \\
        \midrule
        Year & \(4.83 \pm 0.01 \) & \(5.45 \pm 0.01\) & \(5.13 \pm 0.02 \) & \(4.96 \pm 0.01 \) & \(4.93 \pm 0.01\) & \(4.91 \pm 0.02\) & \(\bm{4.680 \pm 0.01}\)\\
        Airline & \(17.90 \pm 0.05\) &  \(17.93 \pm 0.04\) & \(17.53 \pm 0.05\)& \(17.54 \pm 0.07\) & \(17.51 \pm 0.05\) & \(\bm{17.45 \pm 0.04}\)& \(\bm{17.47 \pm 0.03}\)\\
        Taxi & \(283.79 \pm 0.19\) & \(284.22 \pm 0.20\) & \(282.09 \pm 0.32\) & \(\bm{274.65 \pm 0.68}\) & \(281.28 \pm 0.44\) & \(277.60 \pm 0.90\) & \(282.99 \pm 0.21\)\\
        \bottomrule
    \end{tabular}}
    \caption{Negative Log Likelihood and Continuous Ranked Probability Score results on \hyperref[para:large_regression]{large scale regression datasets.}}\label{tab:full_large_regression}
\end{table}

In order to complement the \hyperref[para:co2]{missing values interpolation} results on the \texttt{CO2} dataset. The same experiment is repeated on a Deep GP with a single layer, that is, an sparse GP. The configuration that has been used for all of the experiments is kept here. These results show that a single sparse GP does also suffer from the same problem that Deep GPs. That is, the mean reversion problem when facing a (wide enough) gap in the training dataset.

\begin{figure}[htb!]
\centering
  \includegraphics[width=.7\linewidth]{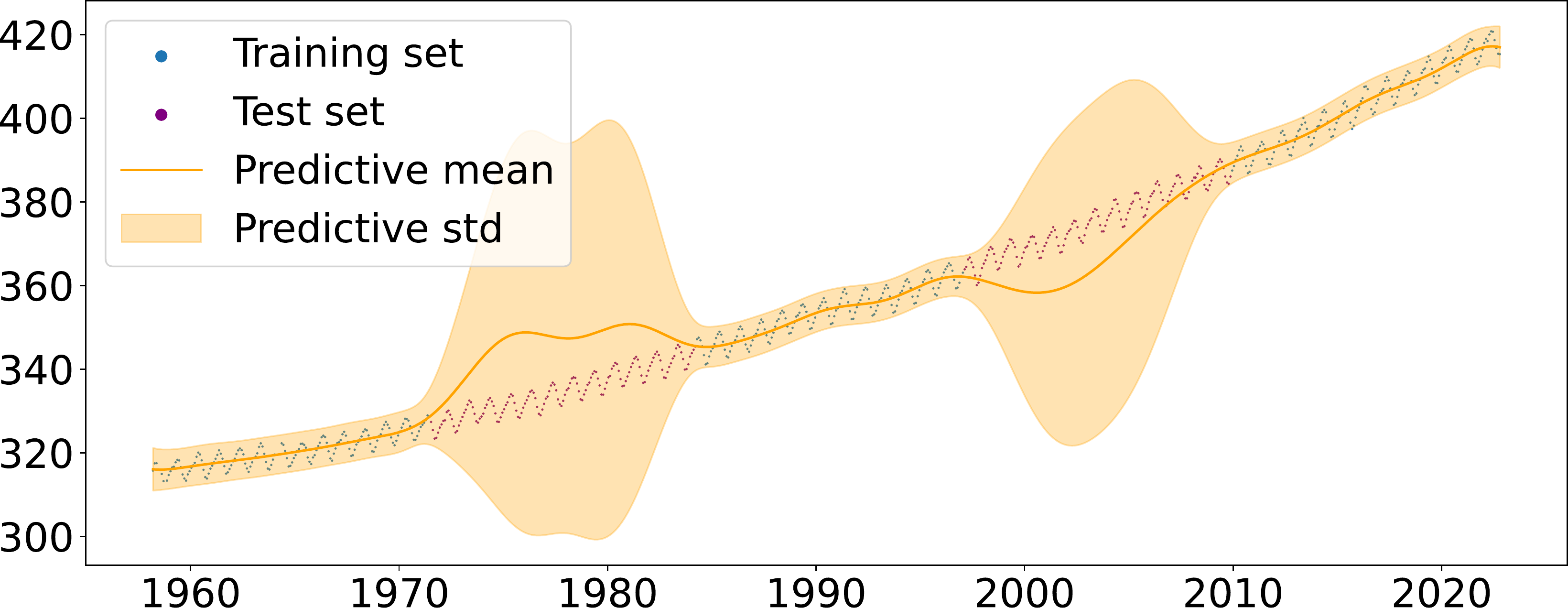}
\caption{\hyperref[para:co2]{Missing values interpolation} results on the \texttt{CO2} dataset. 
Predictive distribution of a sparse GP (Deep GP with a single layer). Two 
times the standard deviation is represented.}
\end{figure}

\section{Potential negative societal impacts}
\label{appendix:sec:potential_neg_impact}

Given that machine learning models are increasingly being used to make decisions that have 
a significant impact on society, industry, and individuals (e.g. autonomous vehicle safety \cite{mcallister2017concrete}, disease detection \cite{sajda2006machine, singh2021better}), it is critical that we have a thorough understanding
of the methods used and can provide rigorous performance guarantees. We conscientiously studied
the performance and application of DVIP to different kinds of datasets and tasks as part of our
empirical evaluation, demonstrating its ability to adjust to each domain-specific dataset.

\begin{landscape}
\begin{table}[b]
    \centering
    \scalebox{0.7}{\begin{tabular}{l*{13}{c}}
        \toprule
        \multirow{ 2.5}{*}{\textbf{NLL}}& \multicolumn{4}{c}{Single-layer} & \multicolumn{4}{c}{Ours} & \multicolumn{4}{c}{\citetalias{salimbeni2017doubly}} \\ \cmidrule(lr){2-5} \cmidrule(lr){6-9} \cmidrule(lr){10-13}
         & SGP & VIP & VIP 200 & SIP & DVIP 2 & DVIP 3 & DVIP 4 & DVIP 5 & DGP 2 & DGP 3 & DGP 4 & DGP 5\\
        \midrule
        \texttt{Boston} &  \(2.62 \pm 0.05\) &  \(2.76 \pm  0.05\)&  \(2.69 \pm 0.03\)& \(2.72 \pm 0.03\) & \(2.85 \pm  0.09\)&  \(\bm{2.59 \pm  0.06}\)&  \(2.67 \pm  0.09\)&  \(2.66 \pm  0.08\)&  \(2.63 \pm  0.05\)&  \(2.63 \pm  0.05\)&  \(2.64 \pm  0.05\)&  \(2.65 \pm  0.05\) \\
        \texttt{Energy}&  \(1.54 \pm  0.02\) &  \(2.07 \pm  0.02\)&  \(2.07 \pm 0.02\)& \(1.17 \pm 0.02\) &\(0.76 \pm  0.02\)&  \(\bm{0.70 \pm  0.01}\)&  \(\bm{0.70 \pm  0.01}\)&  \(0.73 \pm  0.01\)&  \(0.72 \pm  0.01\)&  \(0.74 \pm  0.01\)&  \(0.72 \pm  0.01\)&  \(0.73 \pm  0.01\) \\
        \texttt{Concrete}&  \(3.16 \pm  0.01\) &  \(3.45 \pm  0.02\)&  \(3.48 \pm 0.01\)& \(3.60 \pm 0.05\) & \(3.24 \pm  0.04\)&  \(3.20 \pm  0.05\)&  \(\bm{3.03 \pm  0.02}\)&  \(3.06 \pm  0.02\)&  \(3.17 \pm  0.01\)&  \(3.20 \pm  0.01\)&  \(3.13 \pm 	0.01\)&  \(3.12 \pm  0.01\) \\
        \texttt{Winered}&  \(\bm{0.93 \pm  0.01}\) &  \(0.94 \pm  0.01\)&  \(0.96 \pm 0.01\)& \(0.97 \pm 0.01\) &\(0.94 \pm  0.01\)&  \(0.94 \pm  0.01\)&  \(0.94 \pm  0.01\)&  \(0.95 \pm  0.01\)&  \(0.94 \pm  0.01\)&  \(0.94 \pm 	0.01\)&  \( 0.94 \pm  0.01\)&  \(\bm{0.93 \pm  0.01}\) \\
        \texttt{Power}&  \(2.84 \pm  0.00\) &  \(2.85 \pm  0.00\)&  \(	2.86 \pm 0.00\)& \(2.84 \pm 0.00\) & \(2.82 \pm  0.01\)&  \(2.81 \pm  0.00\)&  \(\bm{2.79 \pm  0.01}\)&   \(\bm{2.79 \pm  0.01}\)& \(2.81 \pm  0.01\)&  \(2.80 \pm  0.00\)&   \(2.80 \pm  0.00\)&  \(2.80 \pm  0.01\) \\
        \texttt{Protein}&  \(2.93 \pm  0.00\) &  \(3.03 \pm  0.00\)&  \(3.03 \pm 0.00\)& \(3.00 \pm 0.00\) & \(2.93 \pm  0.00\)&  \(2.89 \pm  0.00\)&  \(2.88 \pm  0.00\)&  \(2.86 \pm  0.00\)&  \( 2.84 \pm  0.00\)&  \(\bm{2.79 \pm  0.00}\)&  \(\bm{2.79 \pm  0.00}\)&  \(2.80 \pm  0.00\) \\
        \texttt{Naval}&  \(-6.11 \pm  0.06\) &  \(-4.50 \pm  0.02\)&  \(-4.31 \pm 0.00\)& \(-2.79 \pm 0.00\)& \(-5.89 \pm  0.02\)&  \(-5.98 \pm  0.01\)&  \(-5.90 \pm  0.01\)&  \(-5.92 \pm  0.01\)&  \(\bm{-6.35 \pm  0.09}\)&  \(-6.21 \pm  0.04\)&  \(-6.27 \pm  0.06\)&  \(-6.21 \pm  0.08\) \\
        \texttt{Kin8nm}&  \( -0.91 \pm  0.00\) &  \(-0.31 \pm  0.00\)&  \(-0.25 \pm 0.00\)& \(-0.27 \pm 0.02\) &\(-1.00 \pm  0.00\)&  \(-1.13 \pm  0.00\)&  \(-1.15 \pm  0.00\)&  \(-1.16 \pm  0.00\)&  \(-1.29 \pm  0.00\)&  \(-1.32 \pm  0.00\)&  \(\bm{-1.33 \pm  0.00}\)&  \(1.30 \pm  0.00\) \\

        \midrule\midrule
        
        \multirow{ 2.5}{*}{\textbf{RMSE}}& \multicolumn{4}{c}{Single-layer} & \multicolumn{4}{c}{Ours} & \multicolumn{4}{c}{\citetalias{salimbeni2017doubly}} \\ \cmidrule(lr){2-5} \cmidrule(lr){6-9} \cmidrule(lr){10-13}
         & SGP & VIP & VIP 200 & SIP & DVIP 2 & DVIP 3 & DVIP 4 & DVIP 5 & DGP 2 & DGP 3 & DGP 4 & DGP 5\\
        \midrule
        \texttt{Boston} &  \(\bm{3.48 \pm  0.17}\) &  \(4.78 \pm 0.28\)&  \(4.49 \pm	0.28\)& \(5.10 \pm 0.32\)&  \(3.87 \pm	0.19\)&  \(3.50 \pm	0.20\)&  \(3.60 \pm	0.19\)&  \(3.66 \pm	0.21\)&  \(3.51 \pm  0.18\)&  \(3.53 \pm  0.19\)&  \(3.55 \pm  0.20\)&  \(3.56 \pm  0.20\) \\
        \texttt{Energy}&  \(1.07 \pm  0.03\) &  \(2.57 \pm	0.08\)&  \(2.68 \pm	0.07\)& \(3.27 \pm 0.09\)& \(0.52 \pm	0.01\)&  \(0.47 \pm 0.01\)&  \(\bm{0.46 \pm	0.01}\)&  \(0.47 \pm	0.01\)&  \(\bm{0.46 \pm  0.01}\)&  \(0.47 \pm  0.01\)&  \(\bm{0.46 \pm  0.01}\)&  \(\bm{0.46 \pm  0.01}\) \\
        \texttt{Concrete}&  \(5.84 \pm  0.12\) &  \(7.75 \pm	0.15\)&  \(8.06 \pm	0.16\)& \(8.70 \pm 0.43\)  &\(6.01 \pm	0.16\)&  \(5.68 \pm	0.18\)&  \(\bm{5.13 \pm	0.12}\)&  \(5.27 \pm	0.13\)&  \(5.86 \pm  0.12\)&  \(6.01 \pm  0.12\)&  \(5.54 \pm  0.11\)&  \(5.52 \pm  0.12\) \\
        \texttt{Winered}&  \(\bm{0.61 \pm  0.00}\) &  \(0.62 \pm	0.00\)&  \(0.63 \pm	0.00\)& \(0.64 \pm 0.00\)& \(0.62 \pm  0.00\)&  \(0.62 \pm  0.00\)&  \(0.62 \pm  0.00\)&  \(0.62 \pm  0.00\)&  \(0.62 \pm  0.00\)&  \(0.62 \pm 0.00\)&  \( 0.62 \pm  0.00\)&  \(0.62 \pm  0.00\) \\
        \texttt{Power}&  \(4.15 \pm  0.03\) &  \(4.21 \pm	0.03\)&  \(4.22 \pm	0.03\)& \(4.14 \pm 0.03\)  &\(4.06 \pm	0.04\)&  \(4.01 \pm  0.04\)&  \(3.97 \pm  0.04\)&   \(\bm{3.95 \pm  0.04}\)& \(4.00 \pm  0.04\)&  \(3.98 \pm  0.03\)&   \(3.99 \pm  0.03\)&  \(3.96 \pm  0.04\) \\
        \texttt{Protein}&  \(4.56 \pm  0.01\) &  \(5.05 \pm	0.01\)&  \(5.04 \pm	0.01\)& \(4.92 \pm 0.02\)  &\(4.54 \pm  0.01\)&  \(	4.40 \pm  0.01\)&  \(4.33 \pm  0.01\)&  \(4.26 \pm  0.01\)&  \( 4.17 \pm  0.01\)&  \(\bm{4.00 \pm  0.01}\)&  \( 4.01 \pm  0.01\)&  \(4.02 \pm  0.01\) \\
        \texttt{Naval}&  \(\bm{0.00 \pm  0.00}\) &  \(\bm{0.00 \pm  0.00}\)&  \(\bm{0.00 \pm  0.00}\)& \(0.01 \pm 0.00\) &\(\bm{0.00 \pm  0.00}\)&  \(\bm{0.00 \pm  0.00}\)&  \(\bm{0.00 \pm  0.00}\)&  \(\bm{0.00 \pm  0.00}\)&  \(\bm{0.00 \pm  0.00}\)&  \(\bm{0.00 \pm  0.00}\)&  \(\bm{0.00 \pm  0.00}\)&  \(\bm{0.00 \pm  0.00}\) \\
        \texttt{Kin8nm}&  \( 0.09 \pm  0.00\) &  \(0.17 \pm	0.00\)&  \(0.18 \pm	0.00\)& \(0.18 \pm 0.00\) &\(0.08 \pm	0.00\)&  \(0.07 \pm  0.00\)&  \(0.07 \pm  0.00\)&  \(0.07 \pm  0.00\)&  \(\bm{0.06 \pm  0.00}\)&  \(\bm{0.06 \pm  0.00}\)&  \(\bm{0.06 \pm  0.00}\)&  \(\bm{0.06 \pm  0.00}\) \\
        
                \midrule \midrule
        
        \multirow{ 2.5}{*}{\textbf{CRPS}}& \multicolumn{4}{c}{Single-layer} & \multicolumn{4}{c}{Ours} & \multicolumn{4}{c}{\citetalias{salimbeni2017doubly}} \\ \cmidrule(lr){2-5} \cmidrule(lr){6-9} \cmidrule(lr){10-13}
         & SGP & VIP & VIP 200 & SIP & DVIP 2 & DVIP 3 & DVIP 4 & DVIP 5 & DGP 2 & DGP 3 & DGP 4 & DGP 5\\
        \midrule
        \texttt{Boston} &  \(1.79 \pm  0.05\) &  \(2.25 \pm  0.08\)&  \(2.13 \pm  0.08\)& \(2.35 \pm 0.11\) &\(1.91 \pm  .06\)&  \(\bm{1.76 \pm  0.07}\)&  \(1.81 \pm  0.07\)&  \(1.78 \pm  0.06\)&  \(1.79 \pm  0.05\)&  \(1.80 \pm  0.06\)&  \(1.80 \pm  0.06\)&  \(1.81 \pm  0.06\) \\
        \texttt{Energy}&  \(0.62 \pm  0.01\) &  \(	1.27 \pm  0.04\)&  \(1.30 \pm  0.03\)&  \(1.21 \pm 0.04\) &\(0.28 \pm  0.00\)&  \(\bm{0.26 \pm  0.00}\)&  \(\bm{0.26 \pm  0.00}\)&  \(\bm{0.26 \pm  0.00}\)&  \(\bm{0.26 \pm  0.00}\)&  \(\bm{0.26 \pm  0.00}\)&  \(\bm{0.26 \pm  0.00}\)&  \(\bm{0.26 \pm  0.00}\) \\
        \texttt{Concrete}&  \(3.20 \pm  0.05\) &  \(4.29 \pm  0.08\)&  \(4.43 \pm  0.08\)& \(4.69 \pm 0.11\) &\(3.26 \pm  0.07\)&  \(	3.03 \pm  0.09\)&  \(\bm{2.74 \pm  0.05}\)&  \(2.83 \pm  0.05\)&  \(3.21 \pm  0.05\)&  \(3.31 \pm  0.05\)&  \(3.05 \pm  0.05\)&  \(3.04 \pm  0.05\) \\
        \texttt{Winered}&  \(\bm{0.34 \pm  0.00}\) &  \(\bm{0.34 \pm  0.00}\)&  \(0.35 \pm  0.00\)&  \(0.35 \pm 0.00\)&\(\bm{0.34 \pm  0.00}\)&  \(\bm{0.34 \pm  0.00}\)&  \(\bm{0.34 \pm  0.00}\)&  \(\bm{0.34 \pm  0.00}\)&  \(\bm{0.34 \pm  0.00}\)&  \(\bm{0.34 \pm  0.00}\)&  \(\bm{0.34 \pm  0.00}\)&  \(\bm{0.34 \pm  0.00}\) \\
        \texttt{Power}&  \(2.27 \pm  0.01\) &  \(2.31 \pm  0.01\)&  \(2.31 \pm  0.01\)& \(2.27 \pm 0.01\) &\(2.21 \pm  0.01\)&  \(2.18 \pm  0.01\)&  \(\bm{2.14 \pm  0.01}\)&   \(\bm{2.14 \pm  0.01}\)& \(2.17 \pm  0.01\)&  \(2.16 \pm  0.01\)&   \(2.17 \pm  0.01\)&  \(2.15 \pm  0.01\) \\
        \texttt{Protein}&  \(2.56 \pm  0.00\) &  \(2.87 \pm  0.00\)&  \(2.86 \pm  0.01\)& \(2.77 \pm 0.00\) &\(2.54 \pm  0.00\)&  \(	2.43 \pm  0.00\)&  \(	2.38 \pm  0.00\)&  \(2.33 \pm  0.00\)&  \( 2.31 \pm  0.00\)&  \(\bm{2.19 \pm  0.00}\)&  \(\bm{2.19 \pm  0.00}\)&  \(2.20 \pm  0.00\) \\
        \texttt{Naval}&  \(\bm{0.00 \pm  0.00}\) &  \(\bm{0.00 \pm  0.00}\)&  \(\bm{0.00 \pm  0.00}\)& \(\bm{0.00 \pm 0.00}\) &\(\bm{0.00 \pm  0.00}\)&  \(\bm{0.00 \pm  0.00}\)&  \(\bm{0.00 \pm  0.00}\)&  \(\bm{0.00 \pm  0.00}\)&  \(\bm{0.00 \pm  0.00}\)&  \(\bm{0.00 \pm  0.00}\)&  \(\bm{0.00 \pm  0.00}\)&  \(\bm{0.00 \pm  0.00}\) \\
        \texttt{Kin8nm}&  \(0.05 \pm  0.00\) &  \(0.09 \pm  0.0\)&  \(0.10 \pm  0.00\)& \(0.10 \pm 0.00\) &\(0.04 \pm  0.00\)&  \(0.04 \pm  0.00\)&  \(	0.04 \pm  0.00\)&  \(0.04 \pm  0.00\)&  \(	\bm{0.03 \pm  0.00}\)&  \(\bm{0.03 \pm  0.00}\)&  \(\bm{0.03 \pm  0.00}\)&  \(\bm{0.03 \pm  0.00}\) \\
        
        \bottomrule
    \end{tabular}}
    \caption{Negative Log Likelihood, Root Mean Squared Error and Continuous Ranked Probability Score results on  \hyperref[para:uci]{regression UCI benchmark datasets.}}\label{tab:full_uci}
\end{table}
\end{landscape}